%% file: main.tex
\documentclass[10pt,twocolumn,letterpaper]{article}

\usepackage[applications]{wacv}
\input{preamble}

\definecolor{wacvblue}{rgb}{0.21,0.49,0.74}
\usepackage[pagebackref,breaklinks,colorlinks,allcolors=wacvblue]{hyperref}

\usepackage{array}
\usepackage{makecell}
\usepackage{multirow}
\usepackage{amsmath}
\usepackage{amssymb}
\usepackage[dvipsnames]{xcolor}
\usepackage[table]{xcolor}
\usepackage{listings}
\def\wacvPaperID{561} 
\def\confName{WACV}
\def\confYear{2026}

\title{Ego-EXTRA: video-language Egocentric Dataset for EXpert-TRAinee assistance}

\author{Francesco Ragusa\textsuperscript{*}$^{1,2}$,
Michele Mazzamuto\textsuperscript{*}$^{1,2}$, Rosario Forte$^{1}$, Irene D'Ambra$^{1}$,\\ James Fort$^{3}$, Jakob Engel$^{3}$, Antonino Furnari$^{1,2}$, Giovanni Maria Farinella$^{1,2}$\\
\\
$^{1}$Department of Mathematics and Computer Science - University of Catania, Italy\\
$^{2}$Next Vision s.r.l. - Spinoff of the University of Catania, Italy\\
$^{3}$Meta Reality Labs Research, USA
\\
{\tt\small }
}

\begin{document}

\twocolumn[{%
\renewcommand\twocolumn[1][]{#1}%
\maketitle

\begin{center}
    \centering
    \captionsetup{type=figure}
    \includegraphics[width=0.85\textwidth]{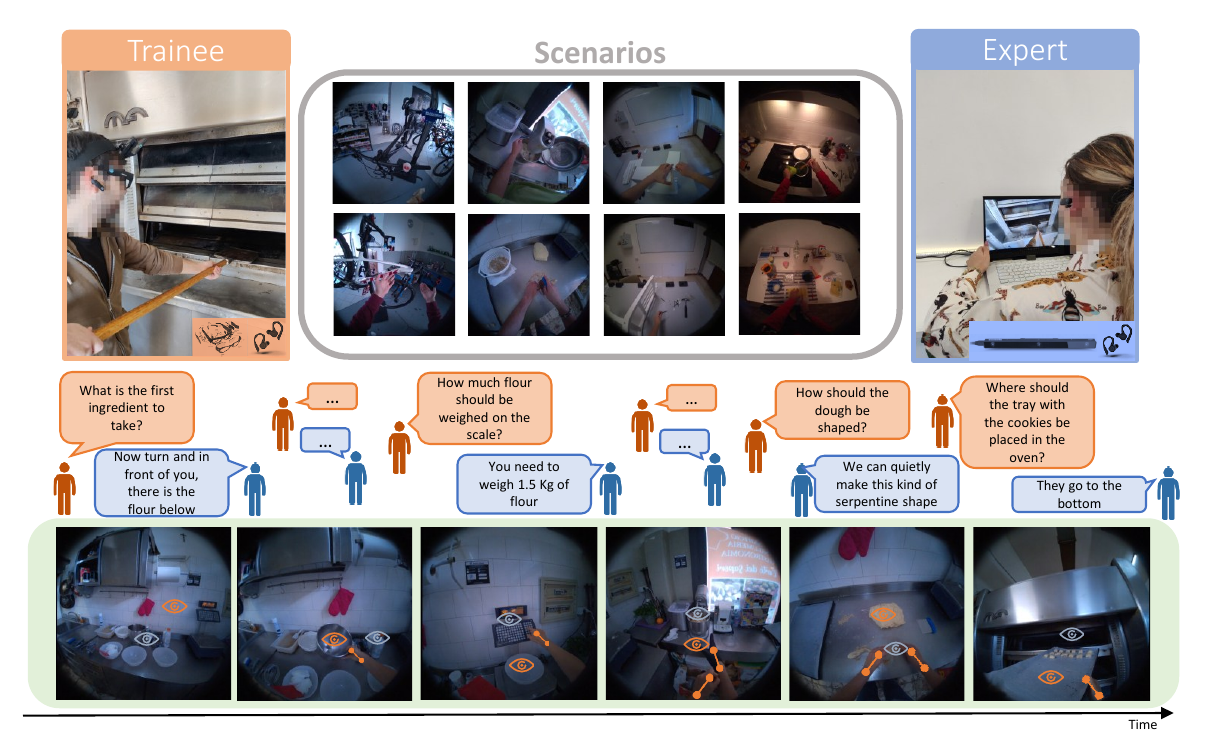}
    \captionof{figure}{We collect egocentric videos of \textcolor{orange}{trainees} (left) performing procedures while aided by an \textcolor{RoyalBlue}{expert} (right) enacting a wearable visual assistant which observes the scene from the trainees' point-of-view and provides guidance through natural language. We gather transcripts of rich natural language dialogue, plus different multimodal signals including eye gaze of both \textcolor{orange}{trainee} (T) and \textcolor{RoyalBlue}{expert} (E), hand keypoints, SLAM, and IMU. The result is a unique set of videos with temporally-aligned dialogue and multimodal signals gathered by Aria glasses.}
\label{fig:concept}
\end{center}%
}]
\begingroup
\renewcommand\thefootnote{}\footnote{\textsuperscript{*}Equal contribution.}
\addtocounter{footnote}{-1}
\endgroup


\input{sec/arxiv_abstract}   

\input{sec/arxiv_introduction}

\input{sec/arxiv_related_work}
\input{sec/3_dataset}

\input{sec/4_benchmark}
\input{sec/6_conclusion}

\input{sec/X_suppl}
\clearpage
{
    \small
    \bibliographystyle{ieeenat_fullname}
    \bibliography{main}
}


\end{document}

%% file: preamble.tex
\renewcommand{\paragraph}[1]{\textbf{#1}}

%% file: sec/arxiv_abstract.tex
\begin{abstract}
We present Ego-EXTRA, a video-language Egocentric Dataset for EXpert-TRAinee assistance. Ego-EXTRA features $50$ hours of unscripted egocentric videos of subjects performing procedural activities (the trainees) while guided by real-world experts who provide guidance and answer specific questions using natural language. Following a ``Wizard of OZ'' data collection paradigm, the expert enacts a wearable intelligent assistant, looking at the activities performed by the trainee exclusively from their egocentric point of view, answering questions when asked by the trainee, or proactively interacting with suggestions during the procedures. 
This unique data collection protocol enables Ego-EXTRA to capture a high-quality dialogue in which expert-level feedback is provided to the trainee.
Two-way dialogues between experts and trainees are recorded, transcribed, and used to create a novel benchmark comprising more than $15k$ high-quality Visual Question Answer sets, which we use to evaluate Multimodal Large Language Models.
The results show that Ego-EXTRA is challenging and highlight the limitations of current models when used to provide expert-level assistance to the user. The Ego-EXTRA dataset is publicly available to support the benchmark of egocentric video-language assistants: \url{https://fpv-iplab.github.io/Ego-EXTRA/}.

\end{abstract}

%% file: sec/arxiv_introduction.tex
\section{Introduction}
\label{sec:intro}
Every day, people naturally engage in different activities, such as washing a car, repairing a bike, assembling new furniture or preparing dinner. Mastering most of these activities requires time, dedication and, often, the guidance of an expert. Think of a college student learning to cook their favorite dish from their father or how to fix a running toilet from their mother. 
Although the web provides plenty of resources to learn such skills autonomously~\cite{wikihow, eHow, youtube}, wearable devices equipped with vision and computation abilities, such as smart glasses, have the potential to act as a \textit{world expert}, providing guidance in a natural way~\cite{plizzari2024outlook}.
Thanks to their ability to look at the world from the privileged egocentric point of view of the user, wearable assistants can relate
vision to language and contextualize questions like “What is this?” or “What should I do now?”. 
Furthermore, wearable assistants should be able to provide practical suggestions that guide the user through the steps of repairing a bike chain and checking that each step has been performed safely and correctly.
Toward this direction, previous works investigated tasks related to procedural video understanding, such as keystep recognition~\cite{ego4d_goalstep,Grauman_2024_CVPR}, mistake detection~\cite{Flaborea_2024_CVPR,seminara2024differentiabletaskgraphlearning,Mazzamuto_2025_CVPR}, planning~\cite{islam2024propose}, procedure understanding~\cite{Grauman_2024_CVPR}, and proficiency estimation~\cite{doughty2018s,Grauman_2024_CVPR}, as well as in tasks related to natural language processing in egocentric vision~\cite{Cheng2024CVPR,Chen2024ARXIV,Mangalam2023EgoSchemaAD}. 
Egocentric vision has also seen a significant increase in the availability of large-scale datasets~\cite{ego4d_goalstep,Grauman_2022_CVPR,Grauman_2024_CVPR,damen2018scaling}. Although these datasets supported the development of procedural video understanding, they do not capture naturalistic dual-agent conversations paired with realistic egocentric visual observations (see Figure~\ref{fig:concept}). Indeed, they typically include textual information obtained through post-acquisition narrations by the camera wearer~\cite{Damen2022RESCALING}, third-party annotators~\cite{Grauman_2022_CVPR}, or experts' commentary~\cite{Grauman_2024_CVPR}, which makes current video source not directly aligned to the objective of evaluating the performance of wearable procedural assistants.

To address this limitation, we present Ego-EXTRA, a new dataset of EXpert-TRAinee interactions aimed at validating video-language models. Ego-EXTRA is composed of $50$ hours of egocentric procedural videos with real trainee-expert conversations recorded during the video acquisition process. 
The dataset has been collected following the “Wizard of OZ” paradigm historically adopted in experimental psychology~\cite{woz_kelley}, linguistics and dialogue state tracking~\cite{mrksic-etal-2017-neural}, where a human simulates a machine interacting with a user.
In our setting, as shown in Figure~\ref{fig:concept}, a \textcolor{orange}{\textit{Trainee}} wears ARIA glasses~\cite{Somasundaram2023ProjectAA} to acquire data while performing a given procedural activity, while an \textcolor{RoyalBlue}{\textit{Expert}} observes the scene from the trainee's point of view through a laptop, providing them with assistance and answering questions. 
We considered four scenarios (i.e., bike workshop, kitchen, bakery, and assembly) where trainees performed different activities (e.g., replacing bike brake pads or cooking a tart), asking questions to the expert whenever they needed help. Thanks to ARIA's rich sensor suite, we simultaneously captured different signals, including RGB, SLAM, eye gaze, IMU, magnetometer, barometer, GPS, BLE, Wi-fi, hand keypoints, and audio enriching the dataset and aligning to previous data collection protocols~\cite{Grauman_2024_CVPR}.
Conversations between trainees and experts are transcribed to text in order to provide language supervision, resulting in a novel set of egocentric videos associated with dual-agent conversations  temporally aligned to videos, a significant departure from current acquisition protocols.

Based on the natural conversations included in Ego-EXTRA, we designed a benchmark of $~15K$ realistic visually grounded question-answer sets (QA sets) and validate the ability of current Multimodal Large Language Models (MLLMs) in supporting the user with natural language supervision. 
To build the benchmark we designed a novel protocol based on the extraction, automatic generation and manual validation/refinement of QA sets which is scalable and applicable to future collection efforts.
We evaluated $4$ state-of-the-art visual-language models on the proposed VQA benchmark and thoroughly examined their limitations. For comparison, we also evaluated a total of $5$ LLMs using only textual input, highlighting that QA sets are based on video content.
Results show that MLLMs achieve an average accuracy ranging from $29.21\%$ to $41.38\%$, demonstrating that the proposed VQA benchmark is challenging for current methods. The dataset will be publicly released to support research in this area.

In sum, the contributions of this work are: 1) we present Ego-EXTRA, a new dataset acquired in realistic scenarios that comprises $50$ hours of egocentric videos and naturalistic trainee-expert conversations; 2) we build a challenging VQA benchmark with a rigorous human-validation step to ensure the high-quality of QA sets. The benchmark is designed as a test set to validate the assistive ability of models; 3) we evaluate different LLMs and MLLMs on the proposed benchmark to assess their performance when answering trainee's questions; 4) we will release the dataset to support the research community in evaluating visual-language models aimed at assisting humans in real-world scenarios.

%% file: sec/arxiv_related_work.tex
\section{Related Work}
\label{sec:rel_work}

\begin{table*}[]
\resizebox{\linewidth}{!}{%
\begin{tabular}{clccccccc}
&\textbf{Name} & \textbf{Settings/Environment} & \textbf{Scenarios} & \textbf{Val\&Test Hours} & \textbf{\begin{tabular}[c]{@{}c@{}}avg. video\\  duration (min)\end{tabular}} & \textbf{\makecell{Expert-Trainee\\Conversations}} & \multicolumn{1}{c}{\textbf{Modalities}} & \textbf{QA/Instruction} \\ 
\cline{2-9}
\multirow{14}{*}{\rotatebox[]{90}{\textbf{Datasets}}} &
EPIC-Kitchens-100~\cite{Damen2022RESCALING} & Cooking / Real & Kitchens & 25.30 & N/A & \textcolor{red}{X} & RGB & \textcolor{red}{X} \\
&CaptainCook4D~\cite{peddi2024captaincook4ddatasetunderstandingerrors} & Cooking / Real & Kitchens & 94.5 & 15.26 & \textcolor{red}{X} & RGB, depth & \textcolor{red}{X} \\
&LEMMA~\cite{lemma_dataset} & House / Real & Kitchens and Living Rooms & 10.8 & 2.00 & \textcolor{red}{X} & RGB, depth & \textcolor{red}{X} \\
&Ego4D~\cite{Grauman_2022_CVPR} & Multi Domain / Real & Multiple scenarios & 288.70 & 24.11 & \textcolor{red}{X} & \makecell{RGB, Audio, 3D environments,\\ stereo, gaze, IMU, multi-view} & \textcolor{red}{X} \\
&Ego-Exo4D~\cite{Grauman_2024_CVPR} & Skilled Activities / Real & \makecell{Soccer, Basketball, Dance, Bouldering,\\  Music, Cooking, Bike Repair, Health Care} & 85.10 & 15.32 & \textcolor{red}{X} & \makecell{RGB, 7-channel audio, IMU, eye gaze, \\ SLAM, 3D environment point clouds, multiview} & \textcolor{red}{X} \\
&MECCANO~\cite{ragusa_MECCANO_2023} & Industrial-like / Lab & Toy Assembly & 3.15 & 20.79 & \textcolor{red}{X} & RGB, depth, gaze & \textcolor{red}{X} \\
&Assembly-101~\cite{sener2022assembly101} & Industrial-like / Lab & Toy Assembly & 66.80 & 7.10 & \textcolor{red}{X} & RGB, multi-view, 3D hand-pose & \textcolor{red}{X} \\
&ENIGMA-51~\cite{ragusa2023enigma51} & Industrial-like / Lab & Electrical Boards Repairing & 10.35 & 26.28 & \textcolor{red}{X} & RGB, 3D models & 200 real instructions \\
&EMQA~\cite{Datta_Dharur_Cartillier_Desai_Khanna_Batra_Parikh_2022} &Indoor Environment / Synthetic & Exploration & N/A & N/A & \textcolor{red}{X} & RGB & 441 synthetic VQA pairs \\
&EgoVQA~\cite{Fan_2019} & Office / Lab & Object Manipulation & 0.65 & 7.5 & \textcolor{red}{X} & RGB & 580 human VQA pairs \\
&HoloAssist~\cite{HoloAssist2023} & Assistive Tasks / Lab & Object Manipulation & 49.80 & 4.47 & \textcolor{red}{X} & \makecell{RGB, depth, head pose, 3D hand pose, \\ eye gaze, audio} & \textcolor{red}{X} \\ 
\cline{2-9}
\multirow{8}{*}{\rotatebox[]{90}{\textbf{VQA Benchmarks}}} &MM-Ego~\cite{Ye2024ARXIV}*\textasciicircum{} & Multi Domain / Real & Multiple scenarios & 2 & 0.2 & \textcolor{red}{X} & RGB & 7026 synthetic VQA pairs \\
&EAGLE~\cite{Bi2024ARXIV}*\textasciicircum{} & Multi Domain / Real & Multiple scenarios & N/A & N/A & \textcolor{red}{X} & RGB & 400K synthetic instructions \\
&ProMQA$^\circ$~\cite{Hasegawa2024ARXIV} & Cooking / Real & Kitchens & 25 & 6.47 & \textcolor{red}{X} & RGB & 401 synthetic VQA pairs \\
&VidEgoThink*~\cite{Chen2024ARXIV} & Multi Domain / Real & Multiple Scenarios & 204 & 2.74 & \textcolor{red}{X} & RGB & 600 synthetic VQA pairs \\
&EgoPlan-Bench*\textasciicircum{}~\cite{chen2023egoplan} & Multi Domain / Real & Multiple Scenarios & N/A & N/A & \textcolor{red}{X} & RGB & 4,939 synthetic VQA\\
&EgoTaskQA$^{\#}$~\cite{Jia_Lei_Zhu_Huang_2022} & House / Real & Kitchens and Living Rooms & N/A & N/A & \textcolor{red}{X} & RGB & 40000 synthetic VQA\\
&EnvQA~\cite{Gao_Wang_Bai_Chen_2021} & House / Synthetic & Kitchens, Living Rooms, Bedrooms, Bathrooms & 38.77 & 0.2 & \textcolor{red}{X} & RGB & 85072 synthetic VQA\\
&ActPlan-1K~\cite{Su2024ARXIV} & House / Synthetic & Kitchens, Living Rooms, Bedrooms, Bathrooms & N/A & N/A & \textcolor{red}{X} & RGB & \textcolor{red}{X}\\
\cline{2-9}
\rotatebox[]{90}{\textbf{D/B}} &Ego-EXTRA & Assistive Procedural Tasks / Real & Bike Workshop, Kitchen, Bakery, Assembly & 50 & 22.78 & \textcolor{ForestGreen}{\checkmark} & \makecell{RGB, SLAM, Trainee eye gaze, Expert eye gaze, \\ IMU, magnetometer, barometer, GPS, \\  BLE, Wi-fi, hand keypoints, and audio} & \makecell{15000 realistic\\ expert-trainee VQA sets} \\ \cline{2-9}
\end{tabular}
}
\caption{
Comparison of Ego-EXTRA (bottom row) with other egocentric datasets (top rows) and benchmarks (middle rows). * indicates a benchmark based on Ego4D~\cite{Grauman_2022_CVPR}, \textasciicircum{} indicates an extension of EPIC-Kitchens~\cite{Damen2022RESCALING}, $^\circ$ refers to an extension of CaptainCook4D~\cite{peddi2024captaincook4ddatasetunderstandingerrors}, and $^{\#}$ indicates an extension of LEMMA~\cite{lemma_dataset}.
}
\label{tab:comparison}
\end{table*}


\noindent
\paragraph{Expert-Level Assistance in Egocentric Vision}
An appealing feature of wearable egocentric systems is their potential to provide expert-level support to human activities~\cite{kanade2012first,plizzari2024outlook}. 
Previous works investigated a plethora of individual tasks aimed to support the development of such systems, notably including temporal action segmentation~\cite{sener2020temporal,zhang2022actionformer,sener2022assembly101}, action anticipation~\cite{girdhar2021anticipative,zhao2023antgpt,mittal2024can}, mistake detection~\cite{HoloAssist2023,seminara2024differentiabletaskgraphlearning,jang2019epic}, procedure understanding~\cite{ashutosh2024video,seminara2024differentiabletaskgraphlearning,zhou2023procedure,Schoonbeek_2024_WACV}, proficiency estimation~\cite{Grauman_2024_CVPR}, and skill determination from video~\cite{doughty2018s,Doughty_2019_CVPR}. 
While these tasks provide essential building blocks to enable the development of assistive systems, an holistic benchmark to support the development and evaluation of methods is missing. 
Towards this direction, we propose Ego-EXTRA, the first dataset of egocentric videos centered around natural vision-language dialogue interactions between experts and trainees aimed to support the evaluation of systems for user assistant in procedural tasks.

\noindent
\paragraph{Language-Based Egocentric Vision Datasets}
Egocentric vision datasets have often included forms of natural language supervision, usually collected after video acquisition~\cite{Damen2022RESCALING,Grauman_2022_CVPR,Grauman_2024_CVPR}.
Datasets of natural conversations of human-object interactions have also been proposed~\cite{northcutt2020egocom}.
Other works included natural language data in the form of procedural instructions~\cite{ragusa2023enigma51,peddi2024captaincook4ddatasetunderstandingerrors,lemma_dataset}.
While providing natural language data at various levels, these previous works did not explicitly aim to collect the specific language of the expert in natural conversations with the user. Notably, HoloAssist~\cite{HoloAssist2023} and Ego-Exo4D~\cite{Grauman_2024_CVPR} recently proposed data collection paradigms aimed at including instructor or expert language respectively.
In particular, HoloAssist~\cite{HoloAssist2023} includes videos of trainees following a given procedure, supported by an instructor who gives them guidance in natural language. The natural language data is used to seed labels for a number of tasks, including action recognition, mistake detection, and intervention prediction, but the raw natural language data is not publicly available. Ego-Exo4D~\cite{Grauman_2024_CVPR} includes videos of subjects with different levels of expertise performing given procedures autonomously. The experts narrate videos after the acquisition, highlighting areas of improvements and good task executions. 
Similarly to HoloAssist, we collect natural conversations between a trainee and a supervisor. While HoloAssist focuses on simple procedures, 
we target real-world procedures such as repairing a bike and making a tart, and recruit real-world experts similar to Ego-Exo4D. Differently from Ego-Exo4D, we aim to collect real dialogue between experts and trainees \textit{during the execution of the task}, with the aim of capturing all nuances of human-assistant dialogue and provide a realistic benchmark for assistive egocentric vision systems.
Table~\ref{tab:comparison} compares Ego-EXTRA (bottom row) with other existing state-of-the-art datasets (top rows).

\noindent
\paragraph{Egocentric Vision-Language Benchmarks}
Owing to the surge in popularity of language models~\cite{vaswani2017attention,touvron2023llama2openfoundation,touvron2023llama,dubey2024llama3herdmodels}, several works proposed benchmarks of egocentric videos based on visual question answering. Typical paradigms for generating high-quality Visual Question Answer (VQA) samples for model evaluation are leveraging synthetic data~\cite{Gao_Wang_Bai_Chen_2021,Su2024ARXIV,Wong_Chen_Wu_Lei_Mao_Gao_Shou_2022}, and augmenting existing datasets of real egocentric videos with human annotations~\cite{Wong_Chen_Wu_Lei_Mao_Gao_Shou_2022}, either with automatic or semi-automatic generation~\cite{Ye2024ARXIV,hasegawa2024promqaquestionansweringdataset,Cheng2024CVPR,Cheng2024ARXIV,chen2023egoplan,Jia_Lei_Zhu_Huang_2022,Bi2024ARXIV,Mangalam2023EgoSchemaAD}.
We follow a similar paradigm to generate a curated VQA dataset from the natural language conversations of Ego-EXTRA with the goal of providing a scalable benchmark for vision-based assistive systems communicating with users using natural language.
Table~\ref{tab:comparison} compares Ego-EXTRA (bottom row) with other existing state-of-the-art VQA benchmarks (middle rows).

\noindent
\paragraph{Multimodal Large Language Models}
Recent works have focused on enhancing Large Language Models (LLMs) to build Multimodal LLMs capable of processing both vision and language data to tackle complex tasks such as question-answering~\cite{Dong2024ARXIV, chen2023egoplan}, open-ended questions~\cite{ataallah2024minigpt4videoadvancingmultimodalllms} and more general tasks~\cite{li2024llavaonevisioneasyvisualtask, Wang2024ARXIV} like visual question answering, document reading, and mathematical reasoning. Additionally, there is significant interest in evaluating the abilities of MLLMs on zero-shot downstream tasks, without retraining on specific target data~\cite{lin2023video, instruct_blip, Chameleon_Team_Chameleon_Mixed-Modal_Early-Fusion_2024}.
We benchmark a set of recent MLLMs on Ego-EXTRA. Our results highlight the limited performance of current MLLMs in assisting humans in realistic scenarios.

%% file: sec/3_dataset.tex
\section{Data Collection}
\label{sec:dataset}

\noindent
\textbf{General Setup}
We collected Ego-EXTRA following the ``Wizard of OZ'' paradigm~\cite{woz_kelley}, historically adopted in the dialogue state tracking literature~\cite{mrksic-etal-2017-neural} to collect realistic conversation turns between users and machine-like systems enacted by humans.
Each session involves two participants: a \textcolor{orange}{trainee} performing a procedural activity, such as assembling a chair, and an \textcolor{RoyalBlue}{expert} who provides guidance and answers questions to ensure correct execution.
The trainee wears a custom rig (see Figure~\ref{fig:concept}-left) consisting of Aria glasses~\cite{Somasundaram2023ProjectAA}, a smartphone positioned to capture a similar viewpoint, and a set of earbuds. The Aria device records multiple signals, including RGB videos, SLAM, eye gaze, IMU, and hand keypoints, while the smartphone and earbuds enable communication with the expert.
The expert, located in a separate room, observes the trainee's actions through a laptop that streams the egocentric video feed and communicates with them via earbuds using natural language.
To further capture expert behavior, the laptop is equipped with a Tobii Pro Fusion Bar~\cite{tobiibar} that records their gaze as they watch the video stream and interact with the trainee.
The bidirectional audio conversation is recorded and later synchronized with the collected egocentric videos. Locating trainee and experts in different physical rooms ensures that 1) the expert perceives the activity solely from the egocentric point of view of the trainee,  and 2) all communication occurs strictly through natural language.

\noindent
\textbf{Session Acquisition Protocols}
As the first of its kind, Ego-EXTRA aims to capture high-quality interactions between trainees and experts, where questions are related to the procedure at hand, rather than to the location of objects or any other elements peculiar to the environment that an expert unfamiliar with the setting could not answer.
To ensure the relevance of collected interactions, we designed two session acquisition protocols, as detailed below.

\begin{figure}
    \centering
    \includegraphics[width=0.49\linewidth,trim=120 120 100 100,clip]{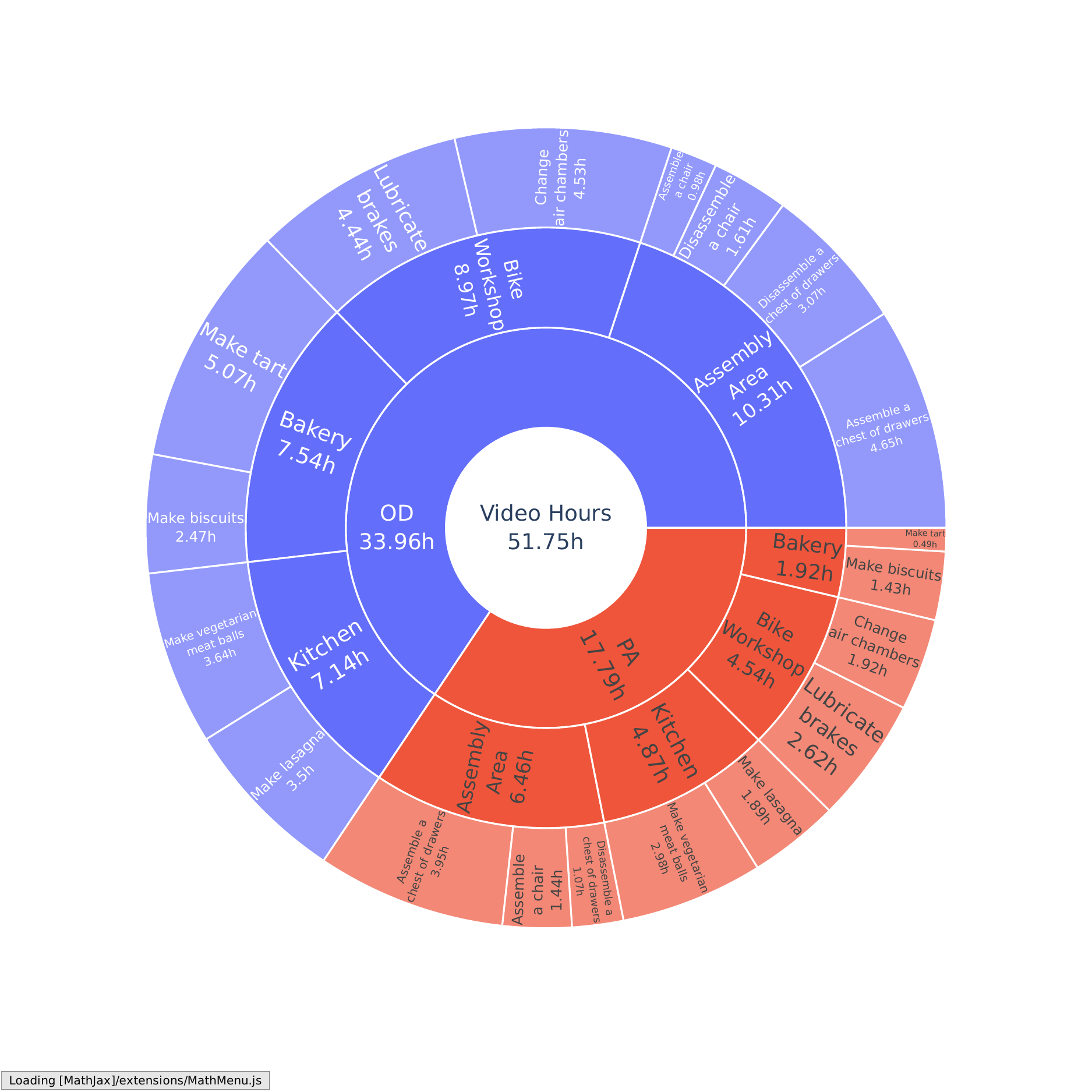}
    \includegraphics[width=0.49\linewidth,trim=120 120 100 100,clip]{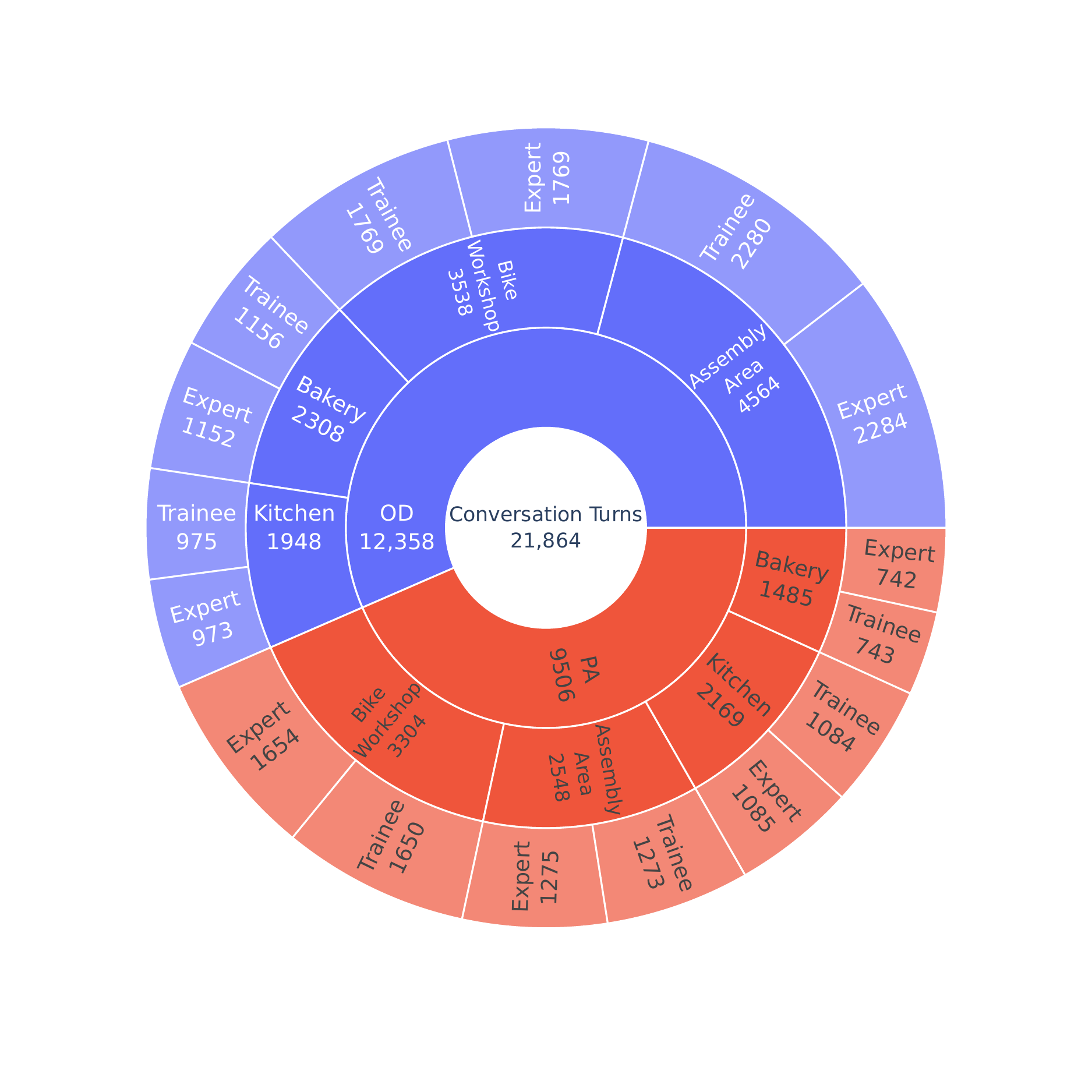}
    \caption{Left: Breakdown of hours per collection protocol, procedure, and scenario. Right: number of expert/trainee conversation turns. PA: Pro-Active, OD: On-Demand.}
    \label{fig:stats}
\end{figure}

\noindent
\textit{Pro-Active protocol (PA).} We instruct the expert to engage in conversations with the trainee in a pro-active way, speaking freely and intervening whenever needed, suggesting next steps, giving instructions, correcting mistakes, and providing any information which is deemed necessary.
A typical intervention of the expert could be: \textit{\textcolor{RoyalBlue}{E: “Firstly, remove the wheel slowly. You should use the wrench that is in the second chest on your left”}}.
Following this protocol, the trainee gets acquainted with the procedure, the environment, the location of objects or tools, and their functions.
This protocol results in dense interactions, with the expert's commentary being predominant and with several conversation turns related to locations of objects and functional areas. Videos have an average duration of 29.64 minutes, with 82.4\% of the words spoken by the expert and an average of 264 conversation turns per video (see Figure~\ref{fig:stats}).

\noindent
\textit{On-Demand protocol (OD). }
We implement this protocol only after the trainee has become familiar with the environment, either by completing a session with the pro-active protocol or by watching a pro-active session conducted by another trainee.
The trainee is hence instructed to carry out the procedure autonomously, interacting with the expert whenever they need guidance, while the expert is instructed to only answer the trainee's questions and to intervene only if a mistake or a potentially dangerous action is about to occur. 
A typical trainee-expert interaction is: \textit{\textcolor{orange}{T: “Which of the two wheels should I remove?”} - \textcolor{RoyalBlue}{E: “The front wheel”}}. 
With this protocol, the dialogue is less dense, but the trainee’s questions are predominant, resulting in a more balanced word distribution between the trainee and the expert, with 61.39\% of the words spoken by the expert and an average of 142 conversation turns per video. The average duration of videos is 23.41 minutes (see Figure~\ref{fig:stats}). 

\noindent
As we are interested in natural interactions, we reduce the number of pro-active sessions to a minimum, roughly resulting in a $1:3$ ratio between pro-active and on-demand videos. When constructing our VQA benchmark, we manually filter out all irrelevant conversation turns from pro-active videos.


\begin{figure*}
\begin{minipage}{0.48\linewidth}
    \centering
    \includegraphics[width=\linewidth]{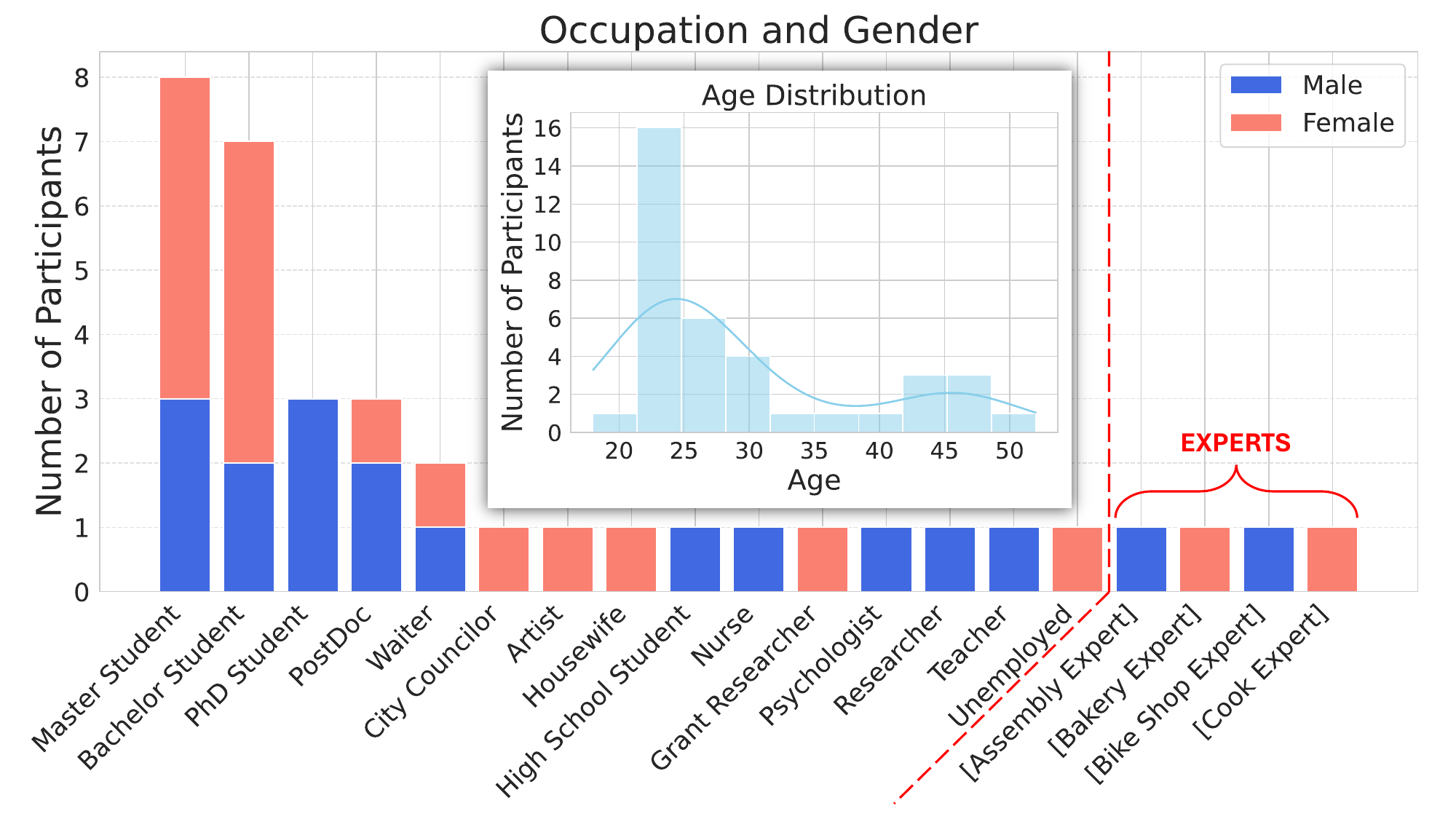}
    \captionof{figure}{Participants' Demographics.}
    \label{fig:demographics}
\end{minipage}
\hfill
\begin{minipage}{0.48\linewidth}
    \centering
    \includegraphics[width=\linewidth]{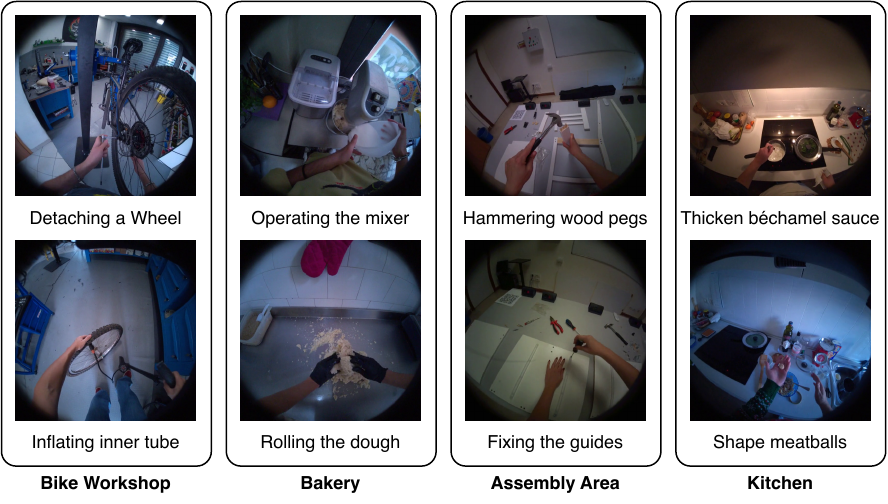}
    \captionof{figure}{Trainees operate in four scenarios, performing varied activities, interacting with objects and tools, assisted by an expert.}
    \label{fig:diversity}
\end{minipage}

\begin{minipage}{0.32\linewidth}
    \centering
    \includegraphics[width=\linewidth,trim=165 0 165 0,clip]{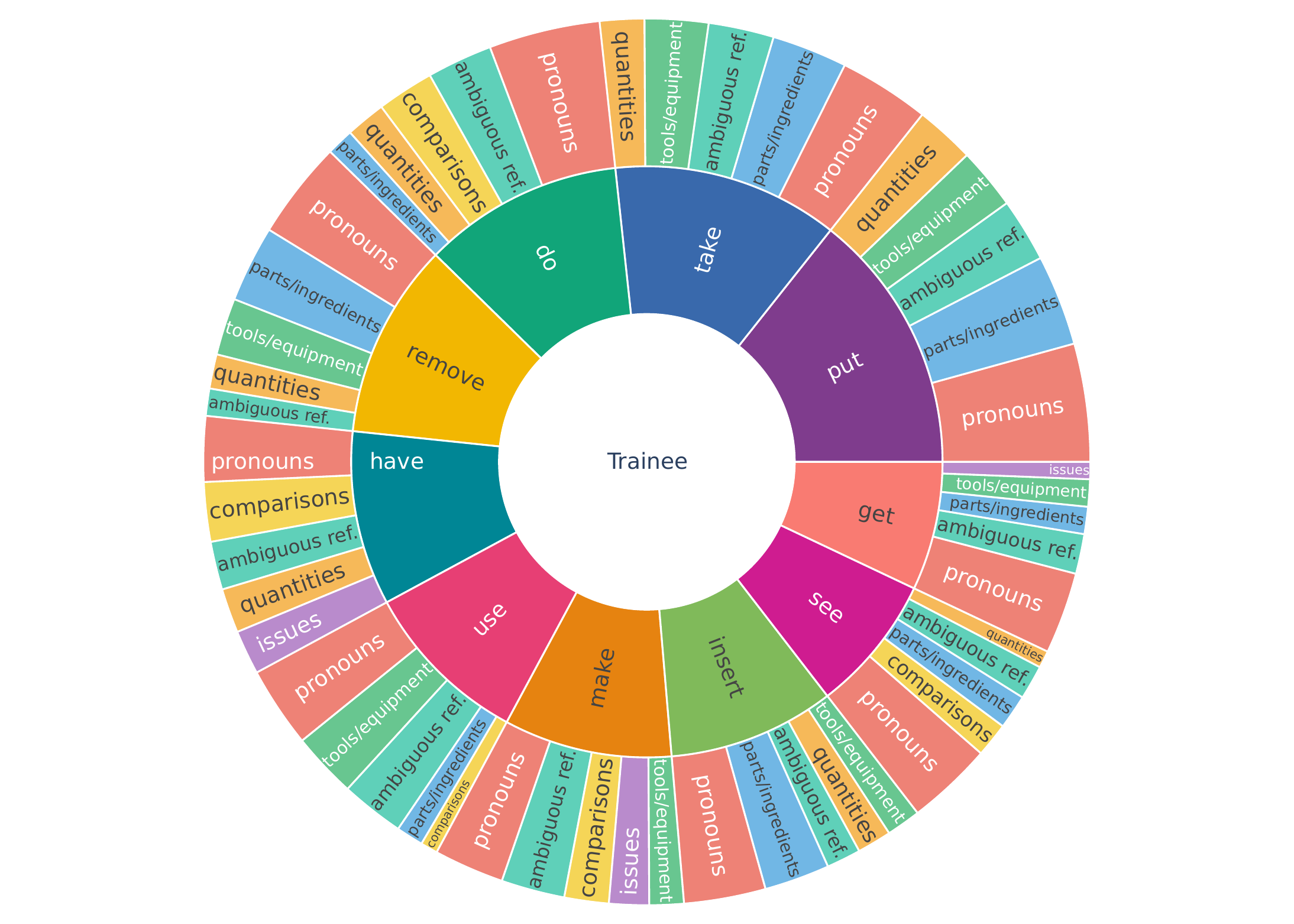}
    \small (a) Trainee Language
\end{minipage}
\hfill
\begin{minipage}{0.32\linewidth}
    \centering
    \includegraphics[width=\linewidth,trim=165 0 165 0,clip]{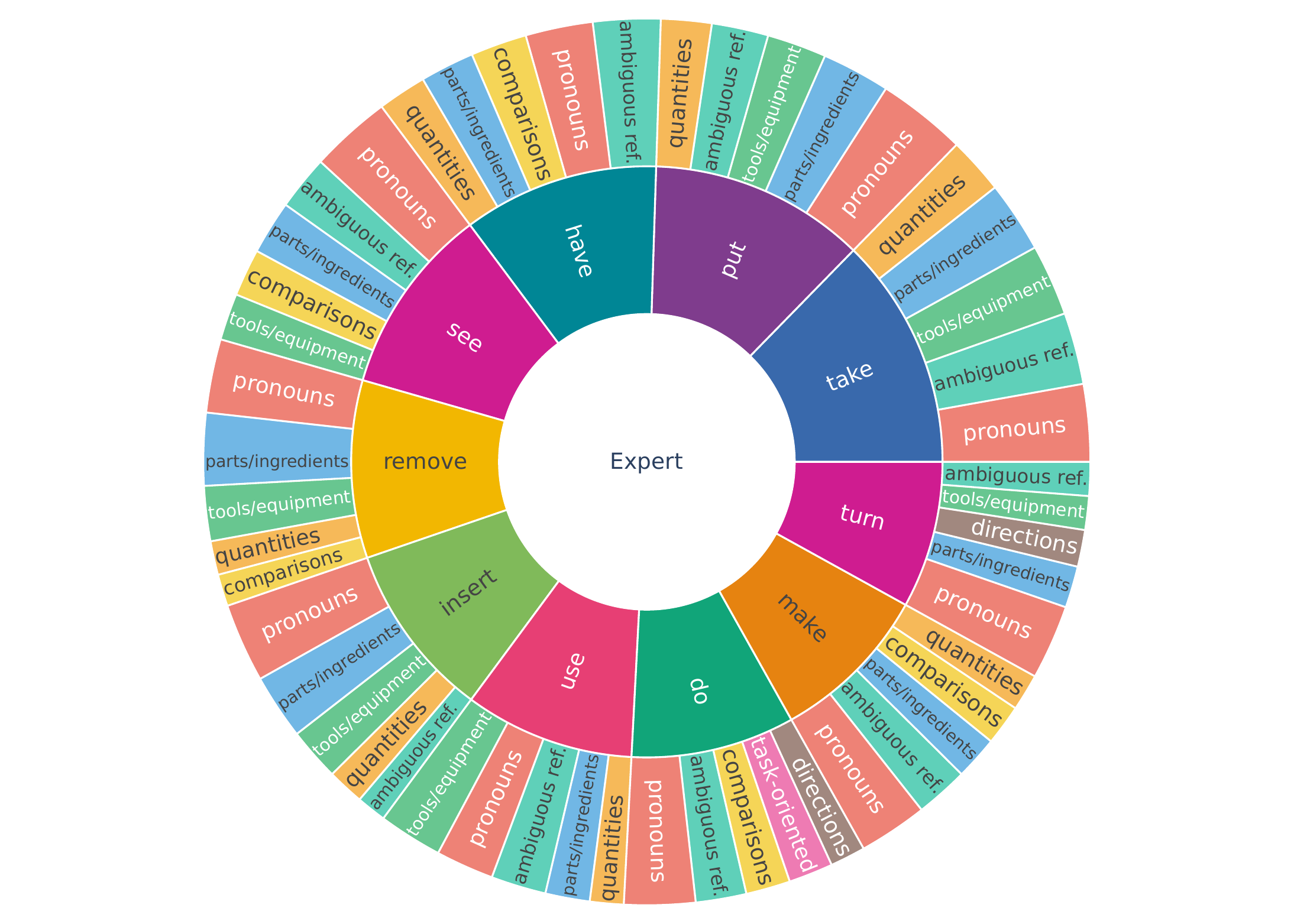}
    \small (b) Expert Language
\end{minipage}
\hfill
\begin{minipage}{0.32\linewidth}
    \centering
    \includegraphics[width=\linewidth,trim=165 0 165 0,clip]{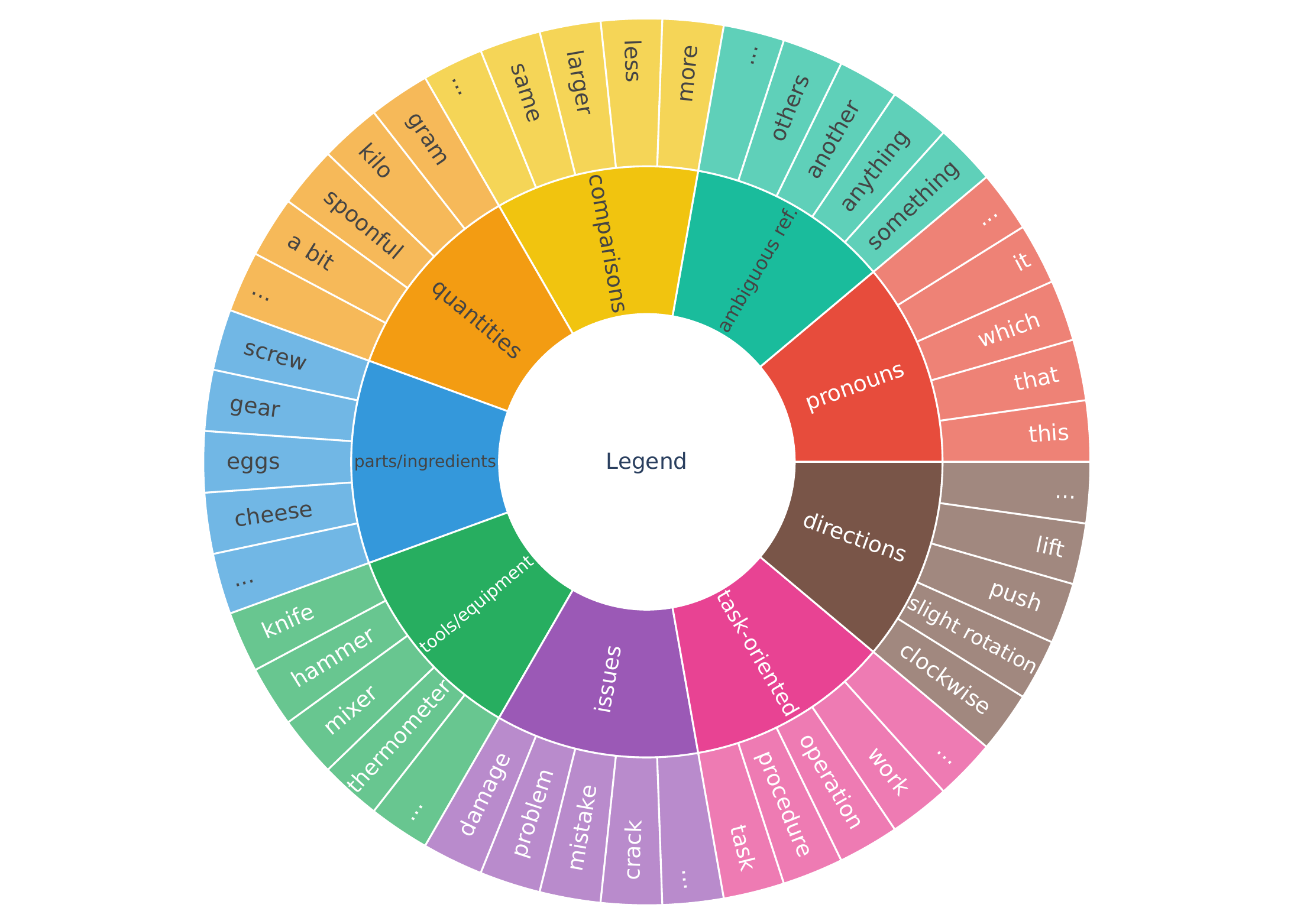}
    \small (c) Legend
\end{minipage}
    \captionof{figure}{Top verb/noun combinations used by trainees (a) and experts (b). Words reported by category, with examples words in (c).}
    \label{fig:language}
\end{figure*}

\noindent
\textbf{Scenarios, Subjects, and Statistics}
We acquired a total of $123$ videos amounting to $50$ hours at $15$ fps with a resolution of 1408x1408 pixels. 
The recorded procedures are split across $10$ different activities and $4$ scenarios, have an average length of $25.24$ minutes, and include a mean of $177.76$ conversation turns per video, about $422.5$ turns per hour or $7$ turns per minute.
See Figure~\ref{fig:stats} for a detailed breakdown of video hours and conversation turns per collection protocol, scenario, and activity. 
Data collection involved a total of 33 trainees and 4 experts across $19$ distinct occupations (see Figure~\ref{fig:demographics}), all volunteers who provided their privacy consent and authorization to acquire data in the considered environments using the described protocols, transcribe audio conversations, and publicly release the resulting data for research purposes. 
Experts were selected as professional figures or individuals with consolidated experience in one of the considered procedures, while we chose trainees with no previous experience in the procedures to ensure realistic interactions. 
While performing the procedures, trainees interact with scenario-specific objects such as an industrial oven in the bakery or inner tubes in the bike workshop, and engage in skilled specialized activities, for which they will require the guidance of the expert (see Figure~\ref{fig:diversity}).

\noindent
\textbf{Postprocessing}
Videos collected through Aria devices are sent to Machine Perception Services~\cite{aria_mps} for the extraction of multimodal signals. 
We then synchronize Aria videos with expert's videos and bidirectional audio conversations.
The expert's gaze is mapped to Aria's viewpoint\footnote{See the supplementary material for the details.}. 
All conversations are transcribed using a professional software and translated from the participants' native language to English using Llama 3.1 405B model~\cite{dubey2024llama3herdmodels}. The quality of the translations has been manually verified. Each conversation turn has a timestamp which allows to localize it in the video.


\noindent
\textbf{Trainee and Expert Language} Figure~\ref{fig:language} (a-b) summarizes the top-10 verbs and top-5 noun category per verb that trainees and experts use in their conversation turns.\footnote{We automatically derive them with NLP techniques.} Figure~\ref{fig:language} (c) reports a legend for noun categories with example nouns. We note a large use of pronouns (``it'', ``this'', ``which''), ambiguous references (``another'', ``something''), and comparisons (``more'', ``less'', ``larger'') which suggests that conversations are naturally grounded in video (also see Figure~\ref{fig:ground_trans}). References to quantities, parts, ingredients, tools, and equipment by both trainees and experts denote a precise language and the need for expert guidance. Language is diversified, with trainees mentioning issues more often than experts, while the latter give directions and use task-oriented language.

\begin{figure*}
\centering
\includegraphics[width=0.8\linewidth]{"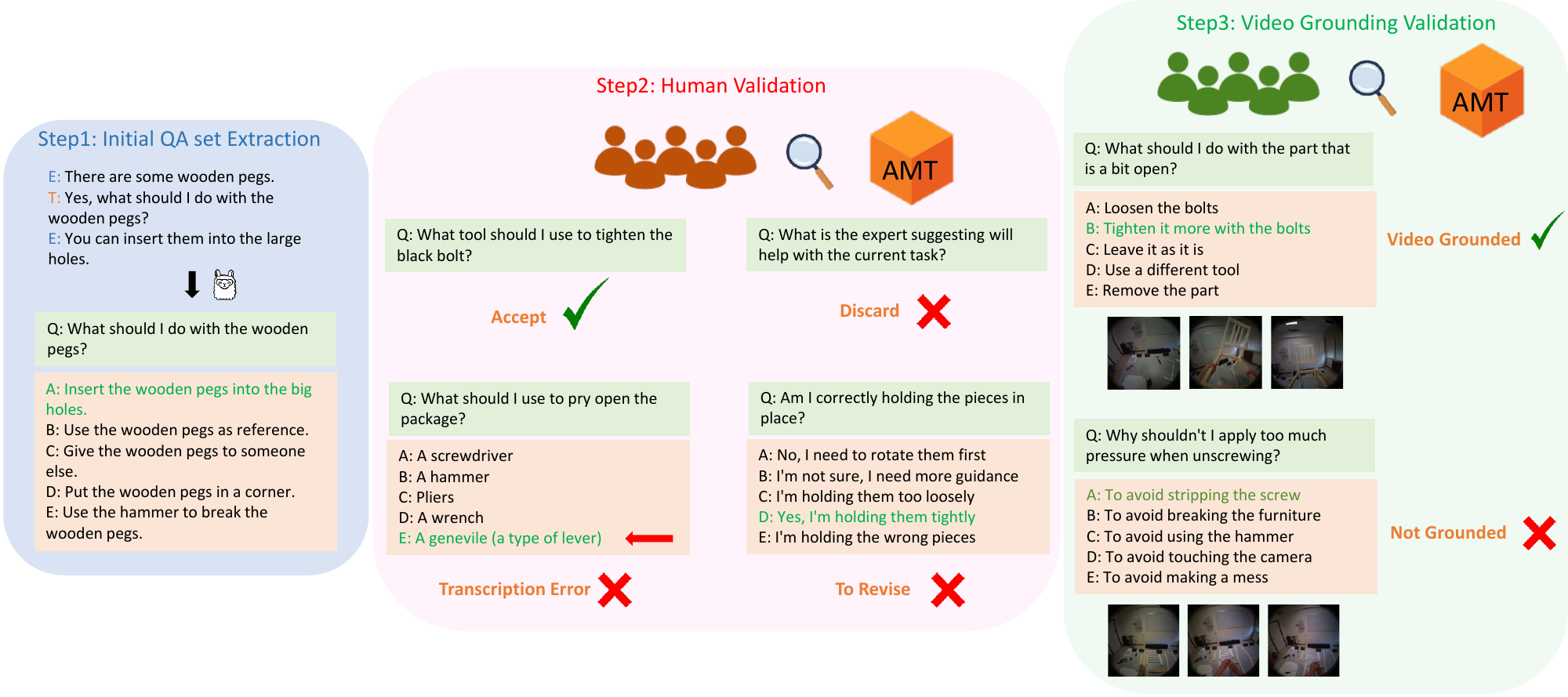"}
\caption{Our multiple-choice question answers generation pipeline: 1) Extraction of QA pairs from transcripts and generation of initial set of negative answers, 2) Human validation, 3) Video Grounding Validation.}
\label{fig:transcript}
\end{figure*}

\section{Ego-EXTRA VQA Benchmark}
Based on the rich vision-language interactions between experts and trainees present in Ego-EXTRA, we built a VQA benchmark designed to evaluate models following the multiple-choice visual question-answering paradigm~\cite{Chen2024ARXIV, Li_2024_CVPR, mm-bench_eccv24, chen2023egoplan}.
In the following, we use the term \textit{question-answer pair (QA pair)} to denote a question and its correct answer, while the term \textit{question-answer set (QA set)} denotes a set made of a question, the correct answer\footnote{Note that each question has only one correct answer.} and four wrong answers, which act as distractors.
We automatically extract QA sets from transcript. Sets are manually checked and manually refined to ensure quality and grounding in video. Figure~\ref{fig:transcript} illustrates the pipeline we follow for QA set generation and validation, while individual steps are discussed in the following.

\begin{figure*}
    \centering
    \includegraphics[width=0.75\linewidth]{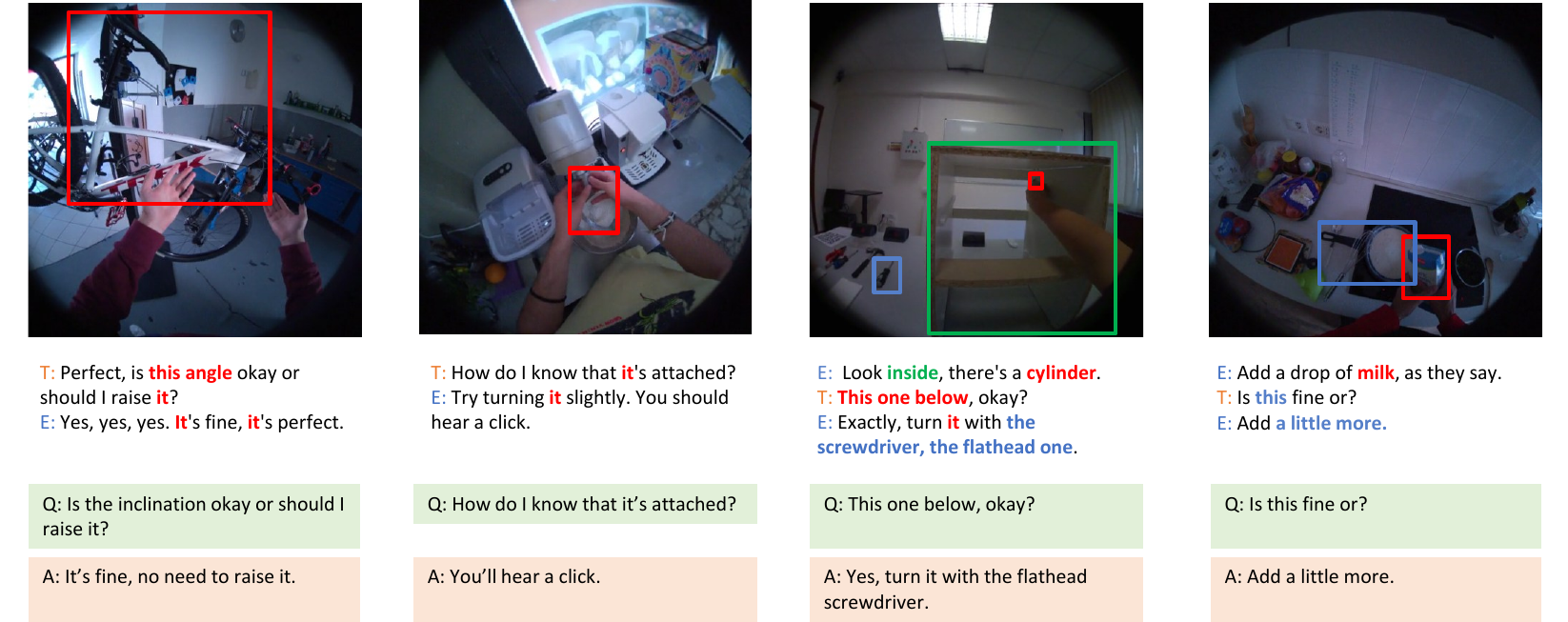}
 \caption{Conversations between trainees and experts are naturally grounded in video since they often refer to objects or their parts. As a result, the generated QA pairs provide a high-quality, visually grounded test-bed for evaluating visual-language models.}
\label{fig:ground_trans}
\end{figure*}

\noindent
\textbf{Step1: Extraction of Initial Question/Answer Sets}
While transcripts contain interactions oriented towards a question-answering scheme, automatically mapping conversation turns to QA sets is not trivial due to the use of filler words, informal exchanges, and unstructured conversations.
We hence resorted to language models for automated analysis. Specifically, we prompted the Llama 3.1 405B model~\cite{dubey2024llama3herdmodels} to 1) extract trainee's questions and expert's answers, 2) automatically correct grammar and transcription errors from the original transcripts, and 3) generate an initial set of $4$ negative (incorrect) answers for each QA pair.
The result is a high-quality set of QA pairs, with initial sets of negatives which will be further refined. While these are generated from textual input, we observe that they are naturally grounded in video thanks to the nature of conversations. 
Figure~\ref{fig:ground_trans} reports examples of question-answer pairs automatically extracted from transcripts, while Figure~\ref{fig:transcript} shows an example of initial QA set generated in step 1.

\noindent
\paragraph{Step 2: Human Validation of QA Sets}
The initial QA sets were manually reviewed by six human validators to filter out irrelevant, generic, ill-formed, or incorrectly transcribed text. We developed a web-based dashboard to support the manual validation process\footnote{See the supplementary material for the details.}. Specifically, we show the question, the correct answer, and a set of negative examples. To facilitate annotators during validation, we also included the conversation turn from which the question was extracted, along with the two preceding and two following turns. As shown in Figure~\ref{fig:transcript}-Step 2, for each QA set, human validators could indicate if the current QA set is \textit{Acceptable}, \textit{To be Discarded}, \textit{Transcription Error} or \textit{Requires Manual Revision}\footnote{In this example, annotators flagged the question for revision because all the answer options were written in the first-person singular.}. Each annotator validated one video for each scenario in Ego-EXTRA. The validation results from this initial set of QAs were then used to design a scalable human validation process on Amazon Mechanical Turk (AMT) to evaluate the entire dataset. Specifically, high-agreement examples from the previous validation phase were used to create a qualification test for AMT workers. Only workers with a global acceptance rate above 90\% and a perfect score (100\%) on the qualification test were allowed to participate. Each QA pair was validated by five independent workers. With this process, approximately 25\% of the initial QAs were discarded.\\
\noindent
\paragraph{Step 3: Video Grounding Validation}
Transcripts of expert–trainee conversations often reference procedural steps and object states that are intrinsically grounded in the video. To ensure this aspect is accurately reflected in the QA sets, we manually reviewed them while also observing the corresponding video clips in which the questions were asked by the trainees (see Figure~\ref{fig:transcript}-Step 3). Following the same pipeline adopted for the Step 2, we provided a dashboard to the six annotators to validate the QA sets providing also the video clip associated to the question. For each question, annotators were asked to select one of the following labels: \textit{Grounded}, \textit{Not Grounded}, or \textit{Video Contains the Answer}. As in the previous phase, we extended the validation process to Amazon Mechanical Turk. Only workers with a global acceptance rate above 90\% and a perfect score on the qualification test were selected. Each QA pair was validated by five independent workers. Using this pipeline, 28\% of the questions were discarded.

\begin{figure}
    \centering
    \includegraphics[width=\linewidth]{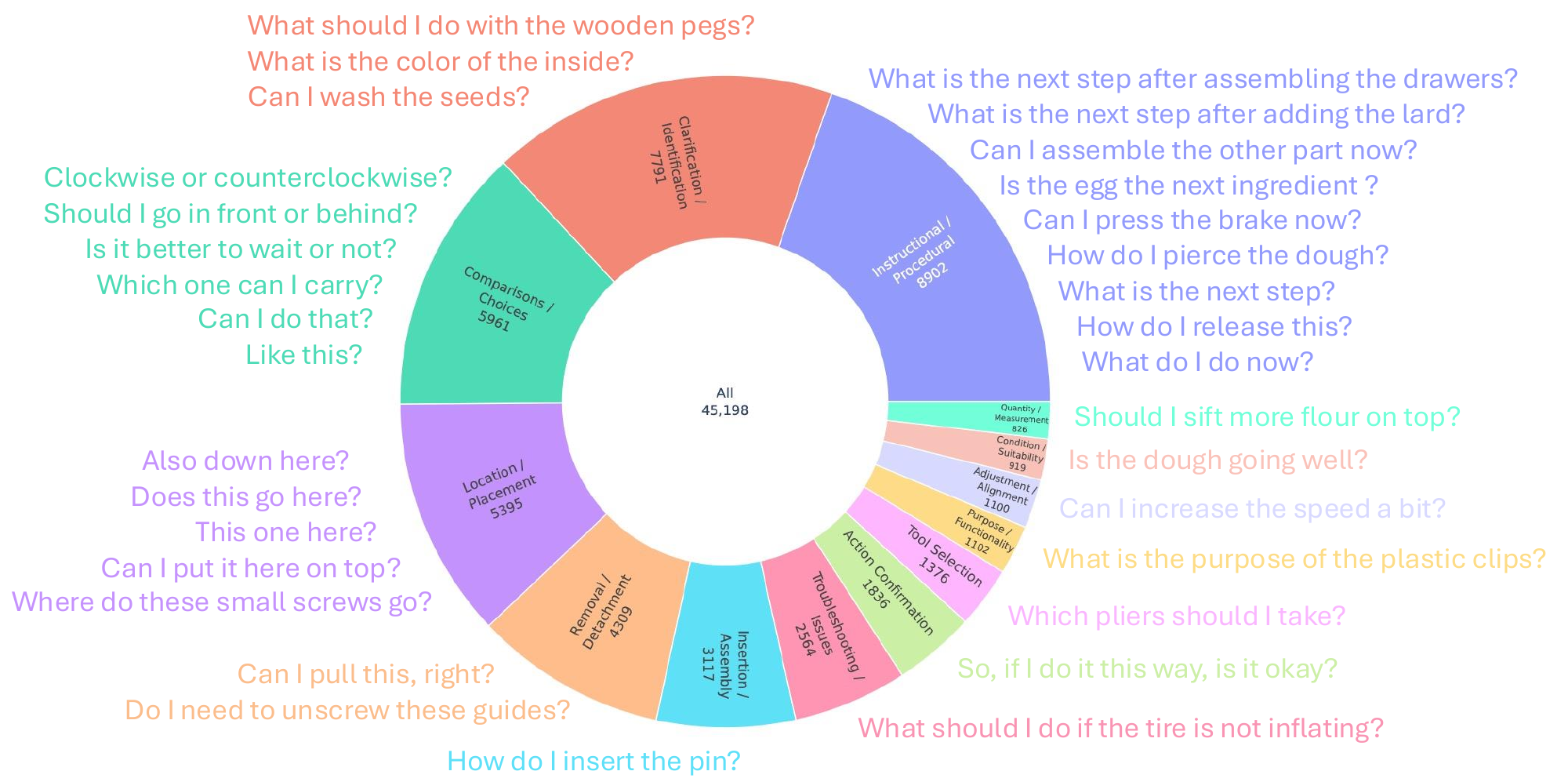}
    \caption{VQA Benchmark Question types and examples.}
    \label{fig:question_types}
\end{figure}

\noindent
\paragraph{Statistics}
We analyze\footnote{We obtain an initial categorization with language models, then we manually refine.} the different types of questions and answers in our benchmarks and identify $13$ main categories, for which we report statistics and examples in Figure~\ref{fig:question_types}.
Note that these questions naturally arise from conversations and do not derive from pre-made templates or any bias introduced during prompting.
The three most prominent question types are Instructional / Procedural (``What do I do now''), clarifications (``What is the color of the inside?''), and comparisons (``Clockwise or counterclockwise?''). 
Specific questions about locations (``Does this go here''), removal (``Can I pull this, right?''), and insertion (``How do I insert the pin), and troubleshooting (``What should I do if the tire is not inflating?'') are also frequent.
Questions about confirmations, tool selection, purpose, alignment, suitability, measurement are overall less frequent, but with a large enough minimum number of instances ($>800$).

%% file: sec/4_benchmark.tex
\section{Experiments}
\label{sec:benchmark}


\noindent
\paragraph{Baselines}
We consider four representative Multimodal Large Language Models (MLLMs) as baselines, namely LLaVA-OneVision~\cite{li2024llavaonevisioneasyvisualtask},  MiniGPT4-video~\cite{ataallah2024minigpt4videoadvancingmultimodalllms}, LLaVa video~\cite{zhang2024videoinstructiontuningsynthetic}, and Qwen2.5-VL~\cite{Qwen2.5-VL}. We also include five LLMs to assess the ability of current models to infer the correct answer purely from text, without any video context. The considered LLMs are Llama 3.3 Instruct Turbo~\cite{llama3.3_instruct_turbo}, Llama 3.1 Instruct~\cite{llama3.1_instruct} 8B and 70B, Qwen 2.5 Instruct 72B~\cite{qwen2.5} and DeepSeek-R1 Turbo~\cite{deepseekai2025deepseekr1}\footnote{More details are in the supplementary materials.}. 
All LLMs are prompted with the QA set and asked to output the correct answer. MLLMs also receive a video clip sampled before the question timestamp ($\delta=5s$). In the main experiments, we feel video models with $8$ frames sampled uniformly from the input clip and ablate on other sampling schemes.
We additionally provide a human baseline on a sample of $217$ questions (an average of $54.25$ per scenario)\footnote{see the supplementary material for more details.}.

\noindent
\paragraph{Benchmark Results}
Table~\ref{tab:vqa_results} reports the overall and per-scenario accuracy of the selected baselines on the VQA benchmark. Language-only models achieve an average accuracy close to a random baseline ($20\%$), ranging from $8.67\%$ with Llama 3.1 Instruct 8B (first row) to $26.65\%$ with Llama 3.1 Instruct 70B (second row).
This highlights that Ego-EXTRA is a challenging benchmark for current LLMs.
Among Video-Language models, LLaVa-OneVision and Qwen 2.5-VL obtain similar performance on average (all around $\sim 30\%$). 
The best performing model is LlaVa-OneVision (third row), which achieves an average accuracy of $33.06\%$, outperforming other models in each scenario, except for the \textit{Kitchen} scenario, where Qwen 2.5-VL (fourth row) obtains the highest accuracy.
Nevertheless, the overall average performance of $33.06\%$ highlights the significant challenge of the proposed VQA benchmark, especially when compared to the human baseline, which achieves an accuracy of $89.65\%$. See the supp. material for qualitative results.

\begin{table}[]
\resizebox{\linewidth}{!}{%
\begin{tabular}{ll|ccccc}
                                & \multicolumn{1}{c|}{\textbf{Model}} & \textbf{Bike Workshop}        & \textbf{Bakery} & \textbf{Assembly} & \textbf{Kitchen} & \textbf{Avg.}  \\ \cline{2-7} 
\multirow{5}{*}{\rotatebox[origin=c]{90}{\textit{\scriptsize \textcolor{gray}{Language Only}}}}  & Llama 3.1 Instruct 8B      & 07.63& 08.62& 07.45& 10.96& 08.67\\
                                & Llama 3.1 Instruct 70B     & 27.57& 22.54& 25.19& 31.30& 26.65\\
                                & Llama 3.3 Instruct Turbo   & 27.14& 18.61& 24.67& 30.42& 25.21\\
                                & Qwen 2.5 Instruct 72B      & 20.27& 15.28& 19.01& 21.54& 19.02\\
                                & DeepSeek-R1 Turbo          & 24.22& 21.94& 21.73& 26.15& 23.51\\ \cline{2-7} 
\multirow{4}{*}{\rotatebox[origin=c]{90}{\textit{\scriptsize \textcolor{gray}{Video-Language}}}} & MiniGPT4-video             & 06.62& 07.09& 08.26& 15.74& 10.68\\
                                & LLaVa Video                & 27.01& {27.16}& {26.12}& 32.09& 28.55\\
                                & LLaVa-OneVision            & \textbf{32.03}& \textbf{33.13}& \textbf{30.88}& \underline{35.77}& \textbf{33.06}\\
                                & Qwen 2.5-VL                & \underline{29.99}& \underline{28.59}& \underline{27.47}& \textbf{35.87}& \underline{31.11}\\ \cline{2-7} 
                                & Sample Human Baseline      & 87.50& 90.91& 100& 81.82& 89.65
\end{tabular}
}
\caption{Results on the proposed VQA benchmark. We report the best results in \textbf{bold} and the second-best results in \underline{underline}.}
\label{tab:vqa_results}
\end{table}

\begin{table}[]
\resizebox{\linewidth}{!}{%
\begin{tabular}{c|c|lllll}
Model & Input & Bike Workshop & Bakery & Assembly & Kitchen & Avg. \\ \hline
\multirow{3}{*}{MiniGPT4-video} & \cellcolor{gray!30}US  & \cellcolor{gray!30}06.62& \cellcolor{gray!30}07.09& \cellcolor{gray!30}08.26& \cellcolor{gray!30}15.74& \cellcolor{gray!30}10.68 \\
 & QA & 06.81 \textcolor{Green}{ \( \uparrow \) 0.19}& 06.44 \textcolor{red}{ \( \downarrow \) 0.65}& 07.95 \textcolor{red}{ \( \downarrow \) 0.31}& 16.10 \textcolor{Green}{ \( \uparrow \) 0.36}& 10.70 \textcolor{Green}{ \( \uparrow \) 0.02}\\
 & TS & 06.00 \textcolor{red}{ \( \downarrow \) 0.62}& 07.42 \textcolor{Green}{ \( \uparrow \) 0.33}& 08.47 \textcolor{Green}{ \( \uparrow \) 0.21}& 15.21 \textcolor{red}{ \( \downarrow \) 0.53}& 10.45 \textcolor{red}{ \( \downarrow \) 0.23}\\ \hline
\multirow{3}{*}{LLaVa Video} & \cellcolor{gray!30}US & \cellcolor{gray!30}27.01& \cellcolor{gray!30}27.16& \cellcolor{gray!30}26.12& \cellcolor{gray!30}32.09& \cellcolor{gray!30}28.55\\
 & QA & 27.62 \textcolor{Green}{ \( \uparrow \) 0.61}& 27.16 \textcolor{Gray}{ \( \pm \) 0.00}& 26.00 \textcolor{red}{ \( \downarrow \) 0.12}& 30.07 \textcolor{red}{ \( \downarrow \) 2.02}& 27.89 \textcolor{red}{ \( \downarrow \) 0.66}\\
 & TS & 26.02 \textcolor{red}{ \( \downarrow \) 0.99}& 23.30 \textcolor{red}{ \( \downarrow \) 3.86}& 24.56 \textcolor{red}{ \( \downarrow \) 1.56}& 29.37 \textcolor{red}{ \( \downarrow \) 2.72}& 26.51 \textcolor{red}{ \( \downarrow \) 2.04}\\ \hline
\multirow{3}{*}{LLaVa-OneVision} & \cellcolor{gray!30}US & \cellcolor{gray!30}32.03& \cellcolor{gray!30}33.13& \cellcolor{gray!30}30.88& \cellcolor{gray!30}35.77& \cellcolor{gray!30}33.06\\
 & QA & 26.58 \textcolor{red}{ \( \downarrow \) 5.45}& 27.99 \textcolor{red}{ \( \downarrow \) 5.14}& 29.41 \textcolor{red}{ \( \downarrow \) 1.47}& 34.67 \textcolor{red}{ \( \downarrow \) 1.10}& 29.56 \textcolor{red}{ \( \downarrow \) 3.50}\\
 & TS & 24.05 \textcolor{red}{ \( \downarrow \) 7.98}& 25.57 \textcolor{red}{ \( \downarrow \) 7.56}& 21.73 \textcolor{red}{ \( \downarrow \) 9.15}& 29.46 \textcolor{red}{ \( \downarrow \) 6.31}& 25.29 \textcolor{red}{ \( \downarrow \) 7.77}\\ \hline
\multirow{3}{*}{Qwen 2.5-VL} & \cellcolor{gray!30}US & \cellcolor{gray!30}29.99& \cellcolor{gray!30}28.59& \cellcolor{gray!30}27.47& \cellcolor{gray!30}35.87& \cellcolor{gray!30}31.11\\
 & QA & 28.40 \textcolor{red}{ \( \downarrow \) 1.59}& 27.23 \textcolor{red}{ \( \downarrow \) 1.36}& 25.90 \textcolor{red}{ \( \downarrow \) 1.57}& 34.09 \textcolor{red}{ \( \downarrow \) 1.78}& 29.48 \textcolor{red}{ \( \downarrow \) 1.63}\\
 & TS & 26.58 \textcolor{red}{ \( \downarrow \) 3.41}& 26.63 \textcolor{red}{ \( \downarrow \) 1.96}& 24.51 \textcolor{red}{ \( \downarrow \) 2.96}& 32.99 \textcolor{red}{ \( \downarrow \) 2.88}& 28.16 \textcolor{red}{ \( \downarrow \) 2.95}
\end{tabular}
}
\caption{Effect of sampling frames from input video. US: sampling 8 frames uniformly. QA: 8 frames before the timestamp of the question. TS: a single frame at the timestamp. \textcolor{Green}{Increments} and \textcolor{red}{decrements} computed considering as reference the \colorbox{lightgray}{US input.}}
\label{tab:input_strategies}
\end{table}

\noindent
\paragraph{Importance of Video Sampling}
Following~\cite{ego-textvqa_angelayao}, we analyze three different strategies for sampling frames from the video clip given as input to the visual-language models: \textit{Uniform Sampling} (sampling $8$ frames uniformly), \textit{QA frames} (the last $8$ frames of the clip), and \textit{TS frame} (a single frame at the question timestamp). 
This last approach also highlights whether image-level understanding is sufficient to solve our VQA benchmark. 
Results in Table~\ref{tab:input_strategies} show that US is leads to best results in average, and  models generally exhibit performance drops with the QA and TS schemes.
Exceptions include the \textit{Bike Workshop} scenario, where MiniGPT4-video and LlaVa Video achieve gains of $0.19\%$ and $0.61\%$, respectively, with the QA scheme; the \textit{Bakery} and \textit{Assembly} scenarios, where MiniGPT4-video gains $0.33\%$ and $0.21\%$ with TS inputs; and the \textit{Kitchen} scenario, where MiniGPT4-video improves its performance by $0.36\%$ using QA.
In general, these results highlight the importance of observing the video clip compared to a single image.
To further assess the importance of video length, we evaluate the performance of the best model (LLaVa-OneVision), taking as input 8 frames uniformly sampled from video clips of 5, 15 and 30 seconds. 
Table~\ref{tab:video_length} shows an average drop (last column) of $4.37\%$ and $3.28\%$ when increasing the temporal spans of video clips to 15 and 30 seconds, respectively. Degradation is observed across all scenarios, where accuracy decreases with longer videos.

\noindent
\paragraph{Effect of Textual Context}
Table~\ref{tab:context_qwen} also compares the best performing model LLaVa-OneVision 7B with its language-only counterpart 
when feeding the models with the textual transcripts related to the 5-second video clip.
This allows to assess the ability of language-only models when they are provided with additional context and simulates an assistant with a basic form of memory of previous conversation turns.
The comparison shows how the model benefits from video to answer questions ($33.06\%$) compared to using only textual information ($16.62\%$) or only the transcript ($29.26\%$). Using video and transcript improves average performance, leading to $36.17\%$ accuracy.

\begin{table}[]
\resizebox{\linewidth}{!}{%
\begin{tabular}{c|c|lllll}
\textbf{Model} & \textbf{Video Length (s)} & \textbf{Bike Workshop} & \textbf{Bakery} & \textbf{Assembly} & \textbf{Kitchen} & \textbf{Avg.} \\ \hline
\multirow{3}{*}{LLaVa-OneVision} & \cellcolor{gray!30}5 & \cellcolor{gray!30}32.03& \cellcolor{gray!30}33.13& \cellcolor{gray!30}30.88& \cellcolor{gray!30}35.77& \cellcolor{gray!30}33.06\\
 & 15 & 26.80 \textcolor{red}{ \( \downarrow \) 5.23}& 28.59 \textcolor{red}{ \( \downarrow \) 4.54}& 25.15 \textcolor{red}{ \( \downarrow \) 5.73}& 33.29 \textcolor{red}{ \( \downarrow \) 2.48}& 28.69 \textcolor{red}{ \( \downarrow \) 4.37}\\
 & 30 & 26.31 \textcolor{red}{ \( \downarrow \) 5.72}& 29.65 \textcolor{red}{ \( \downarrow \) 3.48}& 27.52 \textcolor{red}{ \( \downarrow \) 3.36}& 34.00 \textcolor{red}{ \( \downarrow \) 1.77}& 29.78 \textcolor{red}{ \( \downarrow \) 3.28}
\end{tabular}
}
\caption{Analysis on the effect of input video length uniformly sampled at 8 frames. We selected the best performing model from our VQA benchmark for this analysis. Differences are computed considering as reference the \colorbox{lightgray}{5 seconds as video length}.}
\label{tab:video_length}
\end{table}

\begin{table}[]
\resizebox{\linewidth}{!}{%
\begin{tabular}{l|cccccc}
\multicolumn{1}{c|}{Model} & Input & Bike Workshop & Bakery & Assembly & Kitchen & \multicolumn{1}{l}{Avg.} \\ \hline
LLaVa-OneVision 7B* & QA & 17.45& 14.98& 13.90& 19.20& 16.62\\
LLaVa-OneVision 7B* & QA + Transcript & 24.71& 27.53& 33.90& 27.53& 29.26\\
LLaVa-OneVision 7B & QA + Video & \textbf{32.03}& \textbf{33.13}& 30.88& \textbf{35.77}& 33.06\\
LLaVa-OneVision 7B & QA + Video + Transcript & 31.81& 33.74& \textbf{39.78}& 35.50& \textbf{36.17}\end{tabular}
}
\caption{Comparison of the best performing LLaVa-OneVision 7B with its language-only counterpart and the addition of the transcript as input. * denotes the language-only counterpart, which takes only text as input.}
\label{tab:context_qwen}
\end{table}

%% file: sec/6_conclusion.tex
\section{Conclusion}
\label{sec:conclusion}
In this work, we introduced Ego-EXTRA, a novel egocentric dataset designed to validate intelligent wearable assistants that can provide natural language guidance in real-world scenarios. Through realistic expert-trainee conversations, Ego-EXTRA captures the complexities of procedural tasks across various skill-based domains. Based on real conversations, we designed a VQA benchmark and evaluated a range of LLMs and MLLMs, highlighting the challenging nature of the benchmark, the current limitations of text-based LLMs, and the benefits of contextualized video for MLLMs. Data and benchmark are publicly available to support the community in the validation of wearable assistants able to provide language-based expert-level guidance to users.

\section*{Acknowledgements}
This research is supported by Meta Reality Labs, Next Vision s.r.l. and by the project Future Artificial Intelligence Research (FAIR) – PNRR MUR Cod. PE0000013 - CUP: E63C22001940006.


%% file: sec/X_suppl.tex

\section{Supplementary Material}

\subsection{Subjects}
Data collection was carried out with the participation of 33 trainees and 4 experts aged between $18$ and $52$ years. All participants are volunteers who provided their privacy consent and authorization to acquire data in the considered environments using the described protocols, transcribe audio conversations, and publicly release the resulting data for research purposes. Table~\ref{tab:table_people} reports the list of trainee with information about Gender, Age, and Profession. 

\subsection{Data Acquisition}
\begin{figure}
        \centering
        \includegraphics[width=0.6\linewidth]{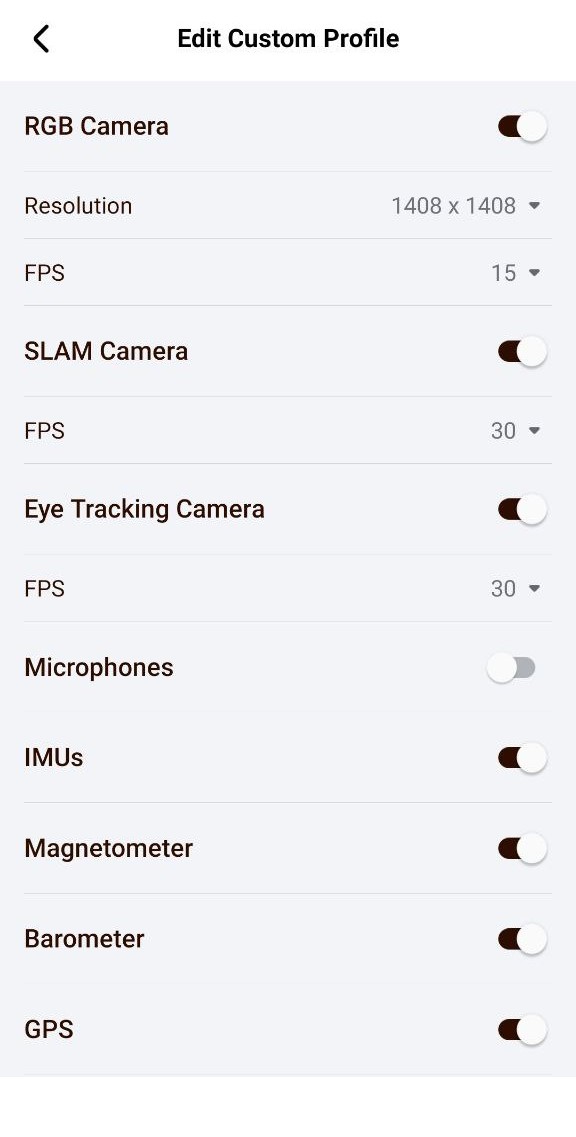}
        \caption{Screenshot of the custom profile used for the acquisition.}
        \label{fig:aria_app}
    \end{figure}

\begin{figure*}[t]
        \centering
        \includegraphics[width=0.99\linewidth]{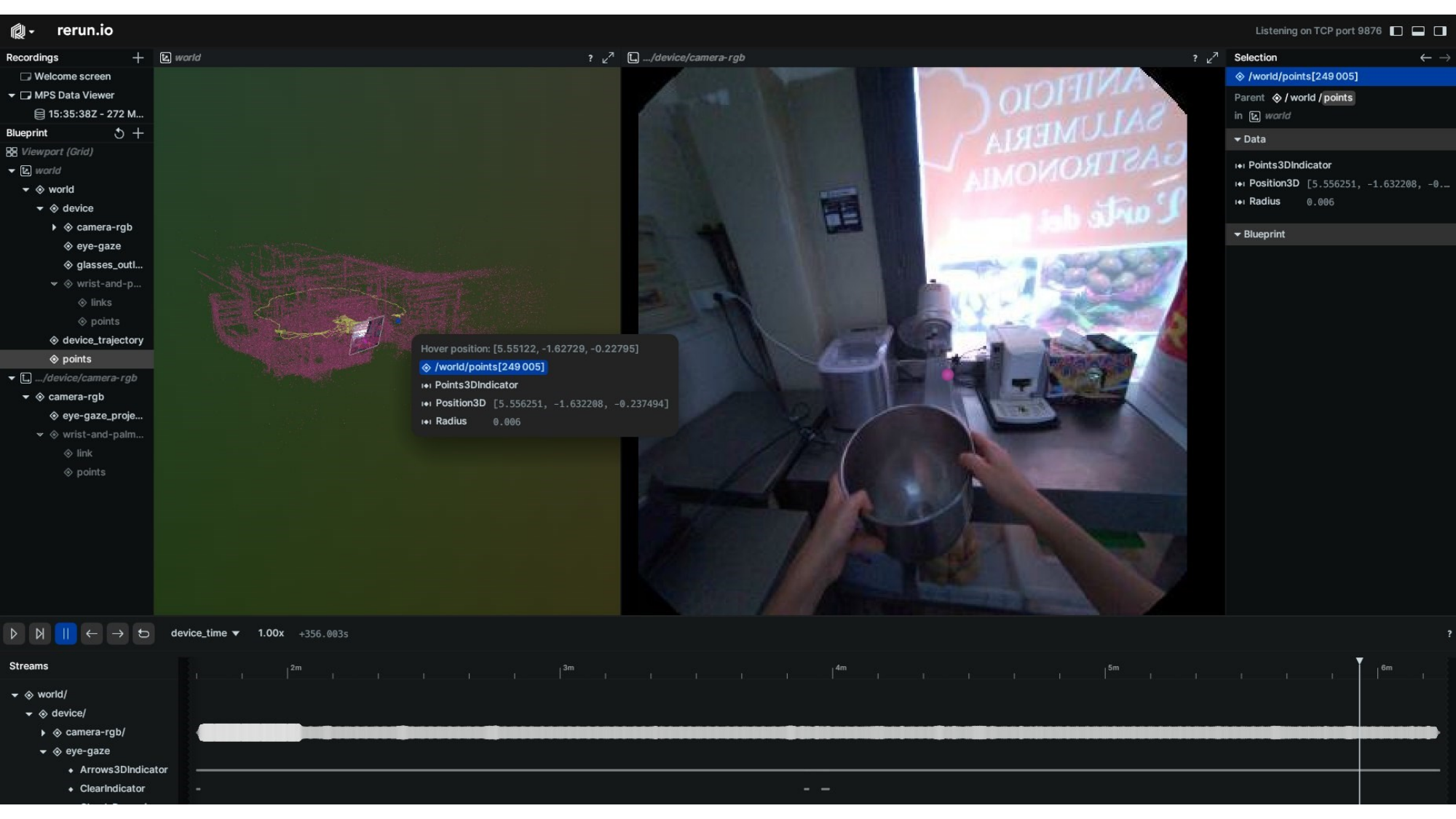}
        \caption{Example of SLAM and eye gaze obtained from the MPS services.}
        \label{fig:mps}
    \end{figure*}

The Aria glasses worn by trainees for data acquisition are equipped with the visual sensors such as  two monochrome scene/SLAM cameras, one RGB camera, and two eye-tracking cameras as well as with non-visual sensors like two inertial measurement units (IMUs), seven-channel spatial microphone array, a magnetometer, a barometer, a GPS receiver, and both Bluetooth and WiFi beacons.
For each acquisition session, Aria glasses are connected to a mobile phone using the ARIA mobile companion app~\cite{aria_companion_app}, allowing the user to manage the data capture process by selecting an acquisition profile. In particular, as shown in Figure~\ref{fig:aria_app} we used a custom profile with the following characteristics:
   \begin{itemize}
       \item RGB camera with a resolution of 1408x1408  at 15 FPS;
       \item SLAM camera at 30 FPS;
       \item Eye-tracking cameras at 30 FPS;
       \item IMUs;
       \item Magnetometer;
       \item Barometer;
       \item GPS;
       \item WiFi and Bluetooth.
   \end{itemize}
The collected data were then exported in VRS (Visual Record Stream) format, which provides standardized methods to store images, audio, and discrete sensor data in compact, evolution-resilient records that are already synchronized.
VRS files are then processed using the ARIA SDK\footnote{\url{https://facebookresearch.github.io/projectaria_tools/docs/ARK/sdk}} to extract the trainee's RGB egocentric video. Synchronized eye gaze and SLAM are obtained using the Project Aria Machine Perception Services\footnote{\url{https://facebookresearch.github.io/projectaria_tools/docs/ARK/mps}} as shown in Figure~\ref{fig:mps}.
Audio conversations have been transcribed using a commercial software. An example of trainee/expert dialogue obtained with both acquisition protocols is reported in Figure~\ref{fig:transcripts}.

\begin{table}[]
\begin{tabular}{|l|l|l|l|}
\hline
Subject Type & Gender & Age & Profession          \\ \hline
Trainee      & M      & 29  & PhD Student         \\ \hline
Trainee      & M      & 23  & PhD Student         \\ \hline
Trainee      & F      & 45  & Grant Researcher    \\ \hline
Trainee      & F      & 36  & Unemployed          \\ \hline
Trainee      & M      & 25  & Master Student      \\ \hline
Trainee      & F      & 25  & Master Student      \\ \hline
Trainee      & M      & 23  & Master Student      \\ \hline
Trainee      & M      & 46  & Teacher             \\ \hline
Trainee      & F      & 30  & PostDoc             \\ \hline
Trainee      & F      & 24  & City Councilor      \\ \hline
Trainee      & F      & 24  & Bachelor Student    \\ \hline
Trainee      & M      & 18  & High School Student \\ \hline
Trainee      & M      & 25  & Waiter              \\ \hline
Trainee      & M      & 52  & Nurse               \\ \hline
Trainee      & M      & 23  & Bachelor Student    \\ \hline
Trainee      & F      & 23  & Master Student      \\ \hline
Trainee      & M      & 33  & Researcher          \\ \hline
Trainee      & M      & 25  & PhD Student         \\ \hline
Trainee      & F      & 22  & Master Student      \\ \hline
Trainee      & F      & 45  & Housewife           \\ \hline
Trainee      & F      & 47  & Artist              \\ \hline
Trainee      & M      & 22  & Bachelor Student    \\ \hline
Trainee      & M      & 28  & PostDoc             \\ \hline
Trainee      & M      & 23  & Master Student      \\ \hline
Trainee      & F      & 23  & Master Student      \\ \hline
Trainee      & F      & 23  & Bachelor Student    \\ \hline
Trainee      & F      & 24  & Bachelor Student    \\ \hline
Trainee      & F      & 24  & Bachelor Student    \\ \hline
Trainee      & M      & 29  & Psychologist        \\ \hline
Trainee      & M      & 28  & PostDoc             \\ \hline
Trainee      & F      & 24  & Bachelor Student    \\ \hline
Trainee      & F      & 24  & Master Student      \\ \hline
Trainee      & F      & 22  & Waiter              \\ \hline
Expert       & M      & 41  & Assembly            \\ \hline
Expert       & F      & 29  & Bakery manager      \\ \hline
Expert       & M      & 47  & Bike Shop Manager   \\ \hline
Expert       & F      & 45  & House Cook          \\ \hline
\end{tabular}
\caption{We reported the list of people engaged in the data acquisition process highlighting their gender, age, and profession.}
\label{tab:table_people}
\end{table}

\begin{figure*}[ht]
    \centering
    \setlength{\fboxsep}{10pt} 
    \setlength{\fboxrule}{1pt} 

    \fbox{
    \begin{minipage}[t]{0.45\textwidth}
    \textbf{Pro-Active:}

    \textcolor{RoyalBlue}{E: Perfect, and the butter will start to melt, and you need to avoid making lumps with the flour, so you need to stir it. I advise you to lower the butter.}\\
    \textcolor{orange}{T: Okay?}\\
    \textcolor{RoyalBlue}{E: Compared to the flour, so make it adhere to the surface of the pan. Okay. Perfect. Wait a moment for it to melt a bit, and set it to four, too. So, the pan doesn't come, doesn't come. Read this signal it's a signal. Move it to the right, move it to the next position, yes.}\\
    \textcolor{orange}{T: Here.}\\
    \textcolor{RoyalBlue}{E: Yes. Put to Four there, or K, perfect. Now, let the butter melt; it will start to melt and we need to mix it with the flour, avoiding any lumps from forming.}\\
    \textcolor{orange}{T: Okay.}\\
    \textcolor{RoyalBlue}{E: Nothing, that little flame was the other burner that turned off. Everything is normal, right? If you think it's too low and the butter is not melting and you want to speed things up, instead of four, turn it to five, you decide, okay?}\\
    \textcolor{orange}{T: Ok I'm setting it to five.}\\
    \textcolor{RoyalBlue}{E: Perfect. It seems that the butter is starting to melt.}
    \end{minipage}%
    \hspace{0.5cm} 
    \vrule width 1pt 
    \hspace{0.5cm} 
    \begin{minipage}[t]{0.45\textwidth}
    \textbf{On-Demand:}

    \textcolor{orange}{T: Since it's already melted for a few seconds, can I leave it? The bechamel sauce.}\\
    \textcolor{RoyalBlue}{E: I advise you to always keep stirring.}\\
    \textcolor{orange}{T: Makes it turn, and then I'll do it, I'll do it with.}\\
    \textcolor{RoyalBlue}{E: You should do, if necessary.}\\
    \textcolor{orange}{T: OK, okay.}\\
    \textcolor{RoyalBlue}{E: Lower the temperature, set it to one, set the bechamel sauce to one and you can leave it. If it's very low, it shouldn't form volumes.}\\
    \textcolor{orange}{T: The cooktop occasionally turns off, so I avoid that by positioning better the pan, right? In the meantime, let's press this.}\\
    \textcolor{RoyalBlue}{E: I see that, that's perfect, good job.}\\
    \textcolor{orange}{T: We help the spinach too?}\\
    \textcolor{RoyalBlue}{E: Wait.}\\
    \textcolor{orange}{T: OK.}\\
    \textcolor{RoyalBlue}{E: Do you know how you can help? By adding a finger of water to that spinach.}\\
    \textcolor{orange}{T: A glass?}\\
    \textcolor{RoyalBlue}{E: Yes, and raise the temperature of the bechamel sauce again if it seems soft.}\\
    \textcolor{orange}{T: Yes, yes, OK, OK. Another minute, precisely, I'll recover the bechamel sauce. Can we drain the spinach?}\\
    \textcolor{RoyalBlue}{E: Yes, but be careful not to burn yourself.}\\
    \textcolor{orange}{T: Is this strainer okay?}\\
    \textcolor{RoyalBlue}{E: You can go.}\\
    \textcolor{orange}{T: Perfect.}
    \end{minipage}
    }

    \caption{Example of trainee/expert conversation acquired with our pro-active (left) and on-demand (right) protocols.}
    \label{fig:transcripts}
\end{figure*}


\subsection{Synchronization and Raw Data Processing}
\subsubsection{Gaze Projection}
\begin{figure}
    \centering
    \includegraphics[width=0.95\linewidth]{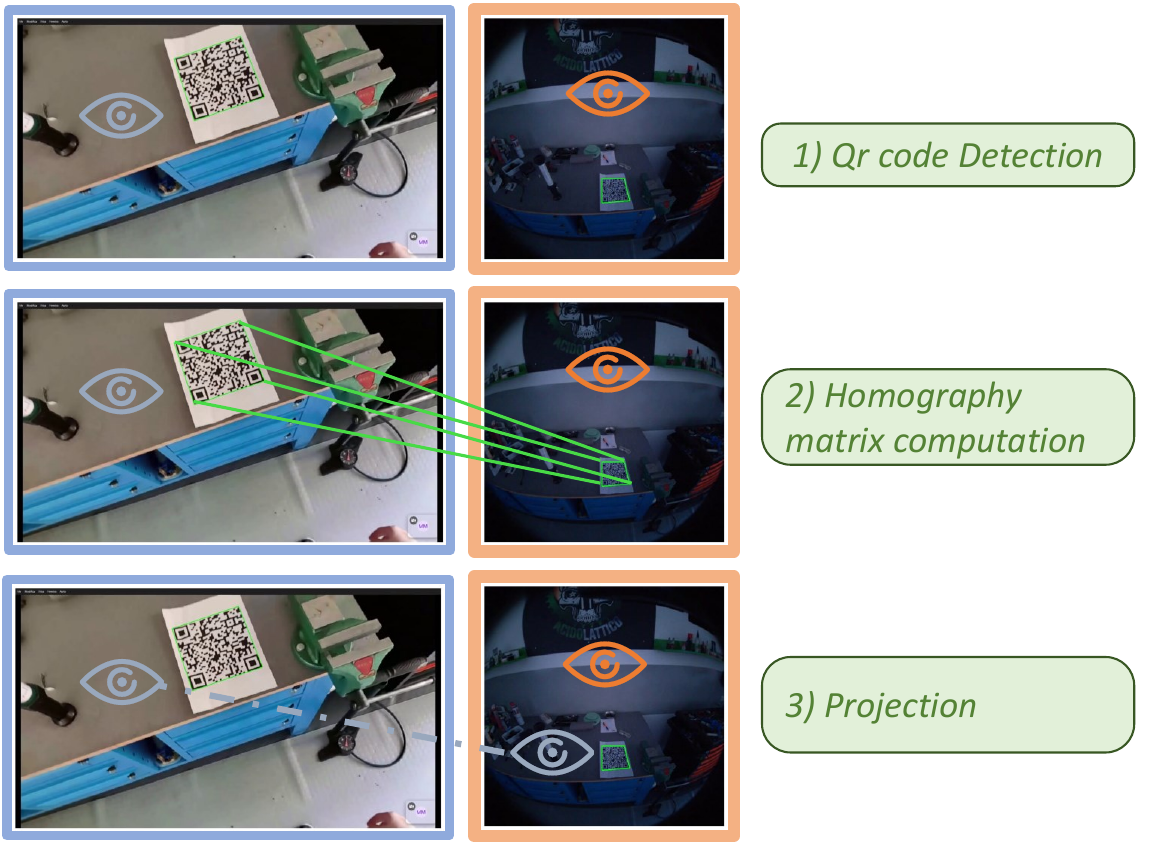}
    \caption{Projection of the expert gaze into the reference point of view of the trainee. In the end of the process, both expert's and trainee's gaze are in the same coordinate system as ARIA's RGB video.}
    \label{fig:gaze_reproj}
\end{figure}

\begin{figure}
    \centering
    \includegraphics[width=1\linewidth]{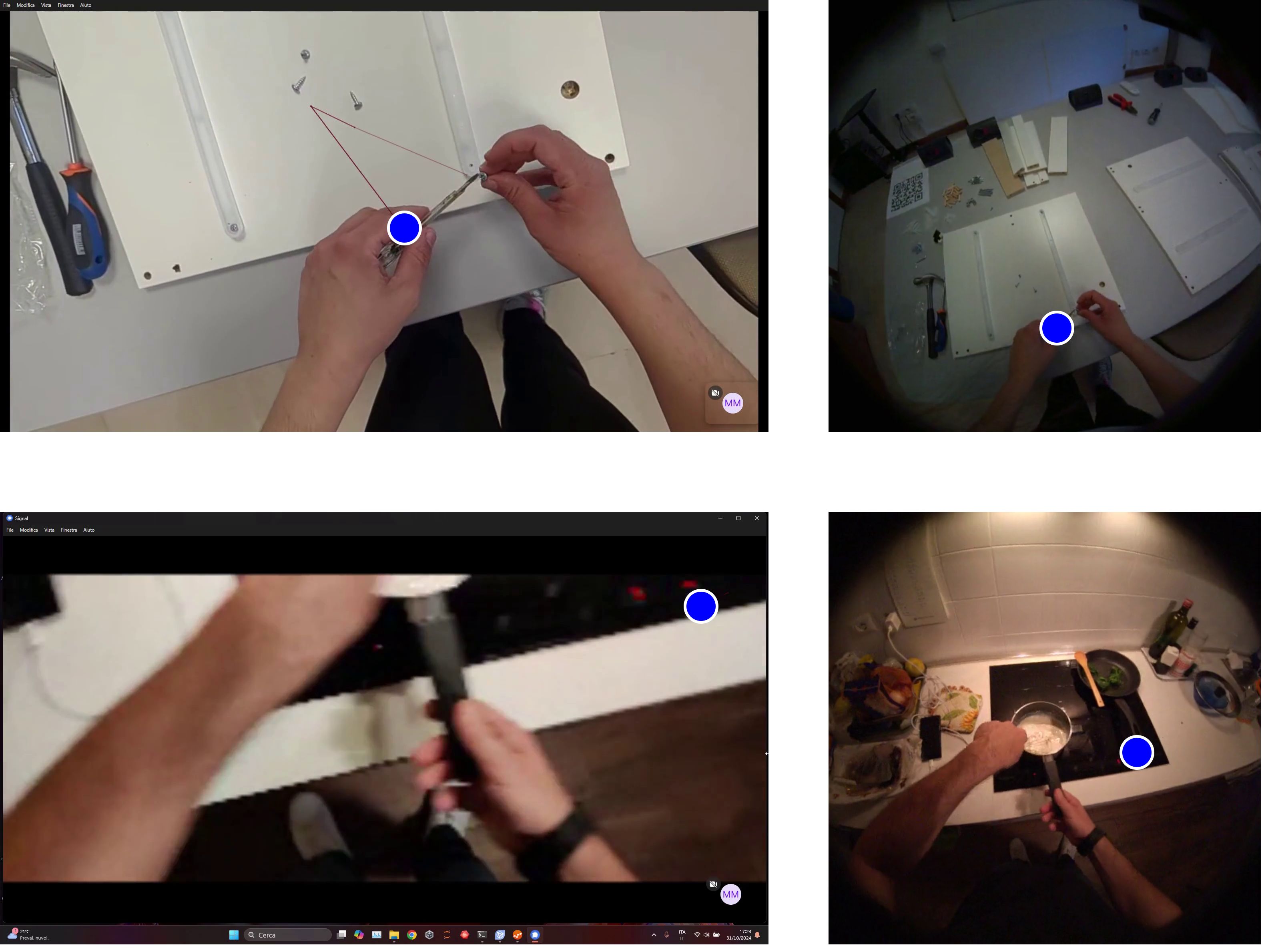}
    \caption{On the left, the trainee's video is streamed to the expert's laptop. On the right, the expert's gaze is reprojected onto the video acquired with the ARIA glasses.}
    \label{fig:gaze_expert}
\end{figure}

To allow a spatial alignment between the egocentric video streams coming from the ARIA device and the smartphone, the trainee was instructed to observe a QR code placed in the environment before starting the acquisition session. The QR code is used to estimate a rigid transformation, allowing the expert's gaze to be projected onto the trainee's viewpoint. The Expert's video stream is recorded together with the gaze signals collected through the Tobii pro device. To allow temporal synchronization between the egocentric video stream and the two-way audio conversation, the trainee and the expert begin each collection by following a countdown to provide a signal useful for temporal synchronization. Based on the recorded countdown, the video pairs are manually synchronized. We detect the QR codes on both the expert's and trainee's videos in the first 60 seconds of each video and compute a \(3 \times 3\) homography matrix \(H\), which stays the same for the duration of the video, that maps the expert's frame to the corresponding trainee's frame. The expert's gaze is therefore projected to the reference frame of the RGB video collected with ARIA, so that both the expert's and trainee's gaze signals are mapped to the same reference system (see Figure~\ref{fig:gaze_reproj} and \ref{fig:gaze_expert}).

\subsubsection{Translation and correction} 
Due to privacy issues, the acquired audio conversations cannot be shared. Therefore, we transcribed all conversations using professional software. We then prompted a LLama 3.1 model to translate the transcriptions into English, correcting any grammar or spelling errors. Each phrase was assigned a timestamp derived from the audio and a unique ID. Table~\ref{tab:transcription_error} reports some examples of corrected transcriptions. 

\begin{table*}[!ht]
\resizebox{\linewidth}{!}{%
\begin{tabular}{cc}
\textbf{Original Transcription} & \textbf{Corrected Transcription} \\ \hline
I have to puttthewater inthe spin acid & I have to put the water in the spinach \\
I mean I don't see lamps & I mean I don't see lumps \\
Is there a specific order in which I have to crew or is it indifferent? & Is there a specific order in which I have to screw or is it indifferent? \\
No, you can pass it withthe clock & No, you can pass it with the cloth \\
That one, thatsilver, this biexactly, thisexam the thisallen key, yes,ok & That one, that silver, this big exactly, this exagonal, this allen key, yes, ok \\ \hline
\end{tabular}
} 
\caption{Some examples of transcription errors that have been corrected.}
\label{tab:transcription_error}
\end{table*}

\section{Ego-EXTRA VQA Benchmark}
\subsubsection{QAs Extraction}
Transforming trainee-expert conversations turns into question-answer sets is challenging. To overcome this issue, we used conversation transcripts to prompt a Llama 3.1 405B model~\cite{dubey2024llama3herdmodels} to generate multiple-choice question answer pairs based on the conversations using a specifically designed prompt reported in the following:
\\

\textit{I will provide you with a transcript of a video. Simulate watching the video and generate questions that can only be answered well if you are watching the video. For each question, generate one correct answer and four incorrect answers (so a total of 5 options). The incorrect answers should be plausible mistakes that could occur during the execution of that action. Avoid trivial questions. Act as a domain expert and generate multiple-choice questions based on the questions asked by (T:) during the provided transcript. Create as many questions as you think are necessary, judging by the length of the transcript and how many questions the apprentice asks (do like from 7 to 15 questions). Each question should include the subject. Never mention the expert or the trainee.}\\

With this prompt the model generates a question, the correct answer and four plausible but incorrect answers as reported in Table~\ref{tab:QA}.


\subsubsection{Human Validation}
In the initial validation phase, six human annotators reviewed the Question-Answering (QA) candidates. Using a dedicated web interface, each annotator was presented with the video clip, conversation transcript, the correct answer, and a set of distractors for each question. Their task was to flag potential issues via checkboxes, including transcription errors, semantically flawed questions, or excessive similarity between the correct answer and the distractors.

To scale up the validation process, we used the results from this initial phase to create a qualification test for Amazon Mechanical Turk (AMT). The test comprised questions that achieved high inter-annotator agreement among our internal team. For the large-scale validation, we selected only AMT workers with a historical approval rate of at least 90

\begin{figure*} \centering \includegraphics[width=0.99\linewidth]{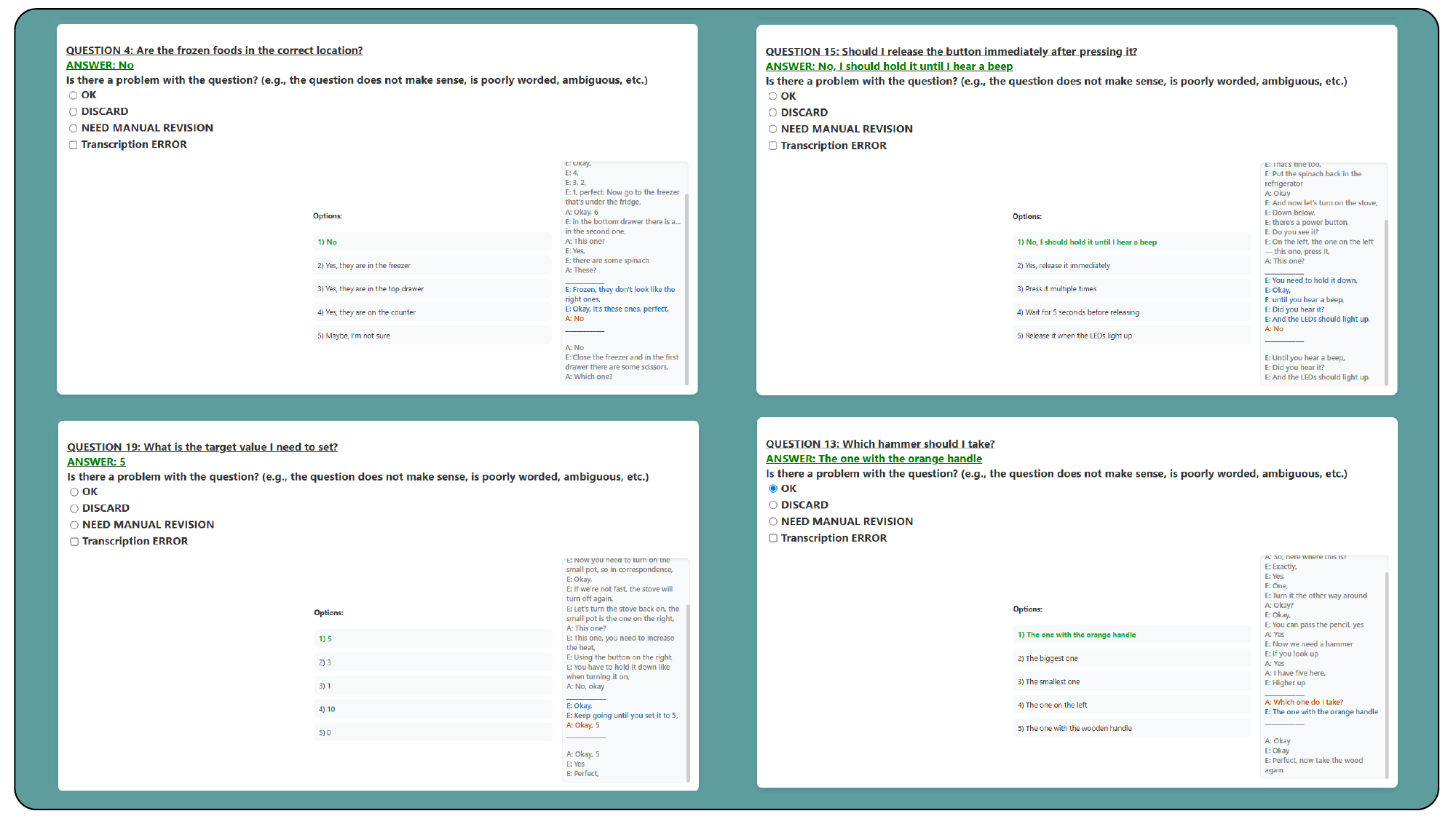} \caption{Tool used for human validation. Each question and option is provided with a checkbox.} \label{fig:tool} \end{figure*}

Examples of questions discarded during this process are reported in Table~\ref{tab:discarded_questions}. The web interface used by the annotators is shown in Figure~\ref{fig:tool}.

\subsubsection{Grounding}  
After the textual validation step, also in this case we perform a two phase grounding validation to ensure that each QA candidate is semantically and visually anchored to the video content. This step is crucial to verify that the question is not only well-formed, but also contextually supported by the video segment and the corresponding dialogue.

Each QA item is manually labeled as \textit{GROUNDED}, \textit{NOT GROUNDED}, or \textit{DISCARD} using a dedicated annotation interface by our six internal annotators. Annotators are presented with the video clip and its transcript, where the current conversation turn is highlighted in color and the surrounding turns are shown in grey for context.

\begin{figure*}
        \includegraphics[width=1\linewidth]{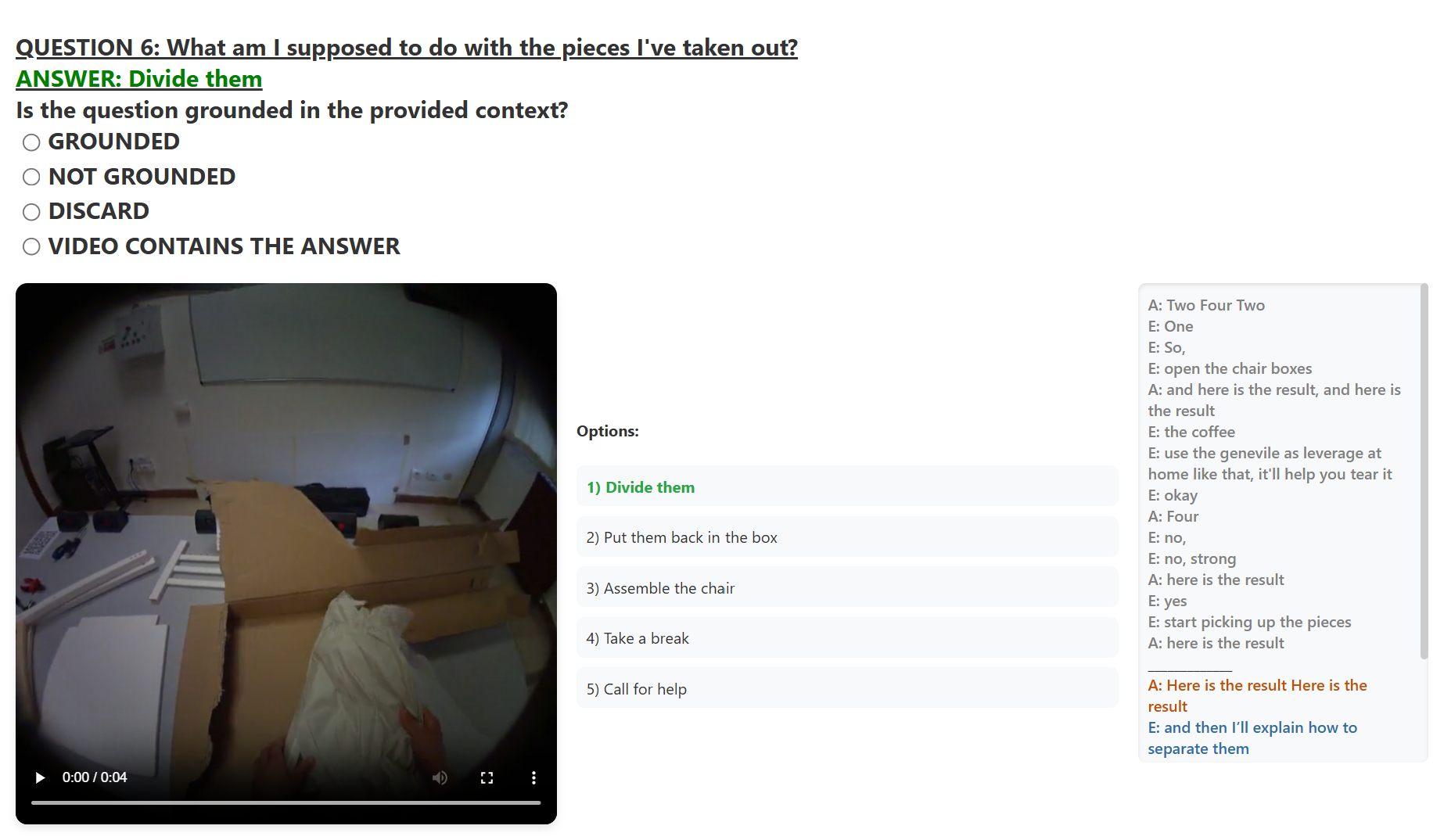}
        \caption{Web tool interface used for grounding validation}
        \label{fig:grounding_tool}
    \end{figure*}

\begin{figure}
    \centering
    \includegraphics[width=1\linewidth]{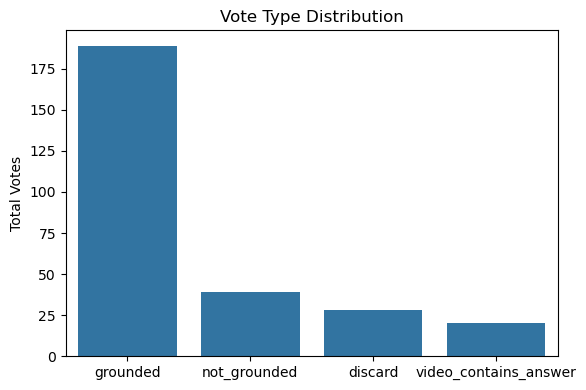}
    \caption{Distribution of grounding labels and average annotation duration.}
    \label{fig:grounding_tool_stats}
\end{figure}

A question is marked as \textit{GROUNDED} if it is clearly supported by the video and coherent with the dialogue. It is labeled as \textit{NOT GROUNDED} if it is unrelated to the visual or textual context, and as \textit{DISCARD} if it is of low quality or not relevant. Annotators also indicate whether the video contains the correct answer to the question. Figure~\ref{fig:grounding_tool} shows the interface used for this task. The distribution of annotations for Scenario 3 across the labeling categories is summarized in Figure~\ref{fig:grounding_tool_stats}. On average, each annotator spent approximately 498.44 seconds completing this task.

Once we obtained labels of grounding for one video per scenario we used them as a qualification set (Similarly to the Human Validation step) to select AMT workers who have an acceptance score above 90\% and a perfect score on the qualification set.

\subsubsection{Baselines}
The following prompt was used for both language-only and video-language models:

\begin{lstlisting}

You are an expert guiding the procedure shown in the video. The question is: '\{question\}'. "Choose the correct answer by selecting one number from the following options:\n" +
            "".join([f"\{i+1\}) \{q\}\n" for i, q in enumerate(options)]) +
            "Reply with ONLY the number of the correct answer (1, 2, 3, 4, or 5). Do not explain or justify. Reply with a SINGLE number."}
\end{lstlisting}
In the case of LLMs, the video is not provided, and they rely solely on the input textual prompt.

\begin{figure*}
    \centering
    \includegraphics[width=1\linewidth]{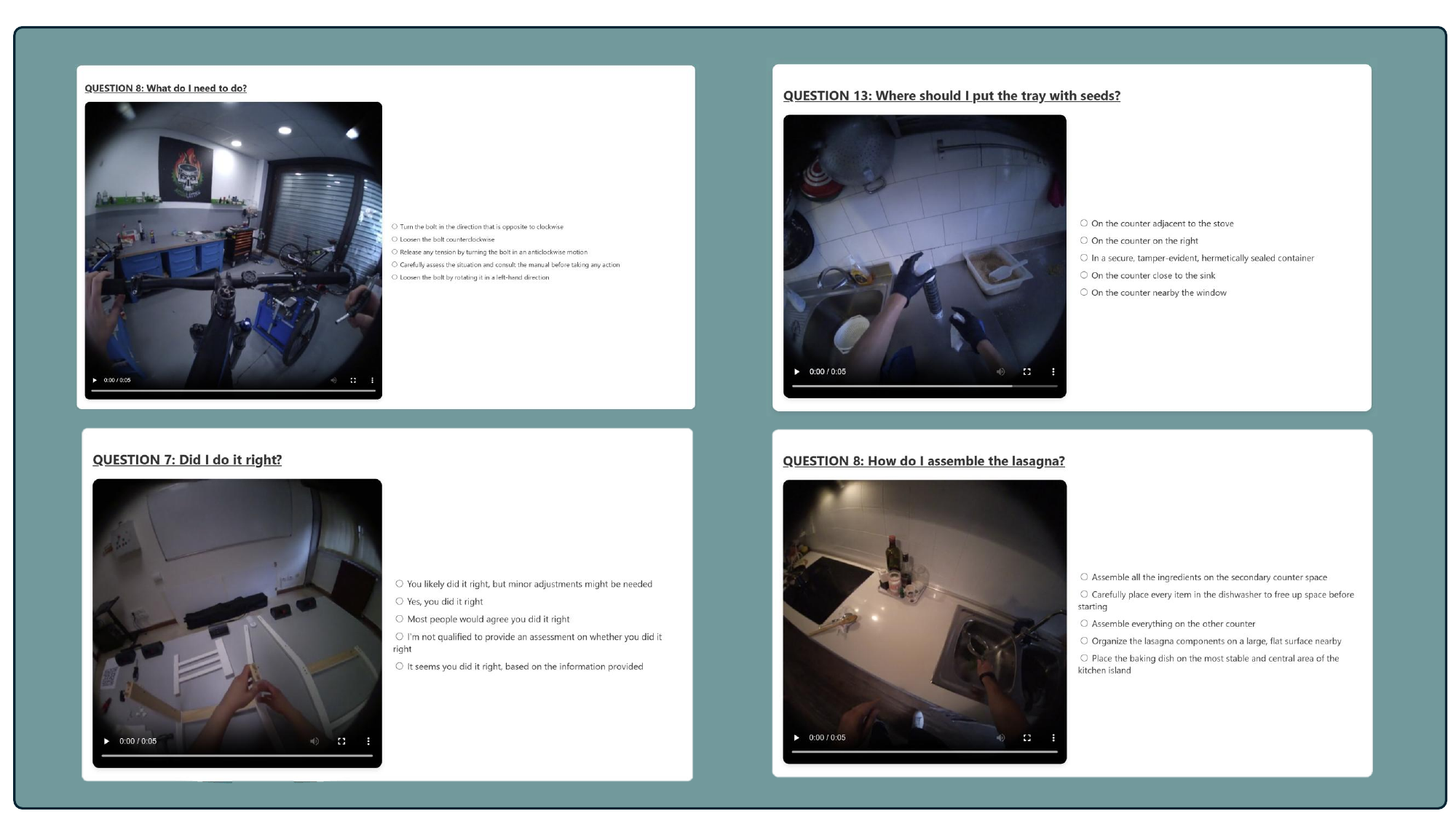}
    \caption{Web tool interface used to obtain the sample human baseline.}
        \label{fig:tool_2_expert}
\end{figure*}

\textbf{Sample Human Baseline}
We provide a human baseline to compare the discrepancy in understanding between humans and state-of-the-art video-language models. We sampled an average of 54.25 questions per scenario, obtaining a total of 217 questions. We designed a web tool to allow experts to answer the questions while observing the related video clips. The experts involved in answering the questions are the same who participated in the data acquisition process. We collected all the answers and computed the human baseline. Example of the web tool interface is shown in Figure~\ref{fig:tool_2_expert}.

\section{Experiments}
\begin{figure*}
    \centering
    \includegraphics[width=1\linewidth]{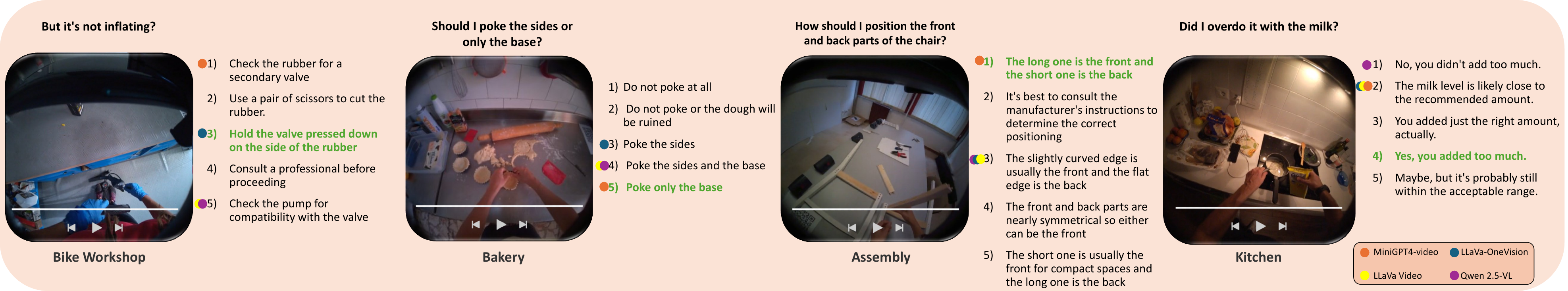}
    \caption{Qualitave results of the proposed VQA benchmark. Correct answer in \textcolor{Green}{green}, baselines predictions marked with colors.}
    \label{fig:qualitative_2}
\end{figure*}

Figure \ref{fig:qualitative_2}-\ref{fig:qualitative} show qualitative results obtained by the adopted baselines in our VQA benchmark.

\begin{table}
\centering
\begin{tabular}{|p{0.5\textwidth}|}
 \hline
 {\textbf{Transcript}}\\
 \hline
 \small
 \textbf{ID 9:} \textcolor{RoyalBlue}{E: Now, let's focus on the next steps.} \newline
 \textbf{ID 10:} \textcolor{orange}{T: Alright, which of the two wheels should I remove first?} \newline
 \textbf{ID 11:} \textcolor{orange}{T: OK, I see.} \newline
 \textbf{ID 12:} \textcolor{RoyalBlue}{E: You should remove the front wheel.} \newline
 \textbf{ID 13:} \textcolor{orange}{T: Great. Is the angle of the bike okay, or should I adjust it?} \newline
 \textbf{ID 14:} \textcolor{orange}{T: OK, understood.} \newline
 \newline \\
 \hline
 {\textbf{QA}}\\
 \hline
 \scriptsize
 \begin{verbatim}
 {
 "id": 1,
 "text": "Which wheel should be removed first?",
 "question_involved_ids": "10-13",
 "options": [
     "The front wheel",
     "The rear wheel",
     "Both wheels",
     "Only the left wheel",
     "Only the right wheel"
 ],
 "correct_answer": "The front wheel",
 "answer_involved_ids": "13",
 "question_start_time": "00:00:09,000",
 "question_end_time": "00:00:15,000",
 "answer_start_time": "00:00:16,000"
 }
 \end{verbatim} \\
 \hline
 \end{tabular}
 \caption{An example of QA generation from the transcript of the trainee/expert conversation.}
 \label{tab:QA}
 \end{table}

\begin{table*}
\centering
\begin{tabular}{|p{0.45\textwidth}|p{0.45\textwidth}|}
 \hline
 \textbf{QA 1} & \textbf{QA 2} \\
 \hline
 \scriptsize
 \begin{verbatim}
"question": "What is the correct way to insert 
the wheel?",
"options": [
    "Insert the wheel from here",
    "Insert the wheel from there",
    "Do not insert the wheel",
    "Insert the wheel with the patches",
    "Insert the wheel without the patches"
],
"correct_answer": "Insert the wheel from here"

 \end{verbatim} & 
 \scriptsize
 \begin{verbatim}
"question": "What is the final state of the chair
after following the instructions?",
    "options": [
        "Assembled",
        "Partially disassembled",
        "Fully disassembled",
        "Broken",
        "Reassembled"
],
"correct_answer": "Fully disassembled"

 \end{verbatim} \\
 \hline
 \textbf{QA 3} & \textbf{QA 4} \\
 \hline
 \scriptsize
 \begin{verbatim}
"question": "What is the purpose of the tare
function in the stand mixer?",
"options": [
    "To measure the weight of the ingredients",
    "To mix the ingredients together",
    "To adjust the speed of the mixer",
    "To reset the mixer to zero",
    "To prepare the mixer for baking"
],
"correct_answer": "To reset the mixer to zero"
  \end{verbatim} & 
 \scriptsize
 \begin{verbatim}
"question": "What is the purpose of crushing
the spinach with a fork?",
"options": [
    "To make the spinach more tender",
    "To make the spinach more flavorful",
    "To help cook the spinach faster",
    "To make the spinach more crunchy",
    "To separate the spinach leaves"
],
"correct_answer": "To separate the spinach leaves"
 \end{verbatim} \\
 \hline
 \end{tabular}
 \caption{Examples of discarded questions by human validation.}
 \label{tab:discarded_questions}
 \end{table*}

    

\begin{table*}[h!]
\centering

\resizebox{\textwidth}{!}{%
\begin{tabular}{|l|l|}
\hline
\textbf{Question Text} & \textbf{Options} \\ \hline
What is the first action to take when disassembling the drawer? & 
\begin{tabular}[t]{@{}l@{}}
1. Pull out the drawer \\ 
2. Remove the plastic clips \\ 
3. Remove the wooden dowels \\ 
4. Unscrew the screws \\ 
5. Use the pliers
\end{tabular} \\ \hline
What is the first action to take when disassembling the drawer? & 
\begin{tabular}[t]{@{}l@{}}
1. Grab the pliers \\ 
2. Pull out the drawer \\ 
3. Remove the screws \\ 
4. Extract the wooden pegs \\ 
5. Remove the plastic clips
\end{tabular} \\ \hline
What is the initial step in taking apart the drawer? & 
\begin{tabular}[t]{@{}l@{}}
1. Pull out the drawer \\ 
2. Loosen the fasteners \\ 
3. Pick up the pliers \\ 
4. Remove the plastic clips \\ 
5. Take out the wooden rods
\end{tabular} \\ \hline
What is the initial step in taking apart the drawer? & 
\begin{tabular}[t]{@{}l@{}}
1. Loosen the fasteners \\ 
2. Pick up the pliers \\ 
3. Remove the plastic clips \\ 
4. Remove the wooden dowels \\ 
5. Open the drawer
\end{tabular} \\ \hline
How do you begin disassembling the drawer? & 
\begin{tabular}[t]{@{}l@{}}
1. Extract the wooden pegs \\ 
2. Remove the plastic clips \\ 
3. Pull out the drawer \\ 
4. Pick up the pliers \\ 
5. Unscrew the screws
\end{tabular} \\ \hline
How do you begin disassembling the drawer? & 
\begin{tabular}[t]{@{}l@{}}
1. Unscrew the screws \\ 
2. Pick up the pliers \\ 
3. Remove the plastic clips \\ 
4. Take out the wooden rods \\ 
5. Pull out the drawer
\end{tabular} \\ \hline
\end{tabular}%
}
\caption{Examples of obtained multiple-choice question answers considering their variants.}
\label{tab:question_variations_full}
\end{table*}

\subsection{Qualitative Results}
\begin{figure*}
    \centering
    \includegraphics[width=\linewidth]{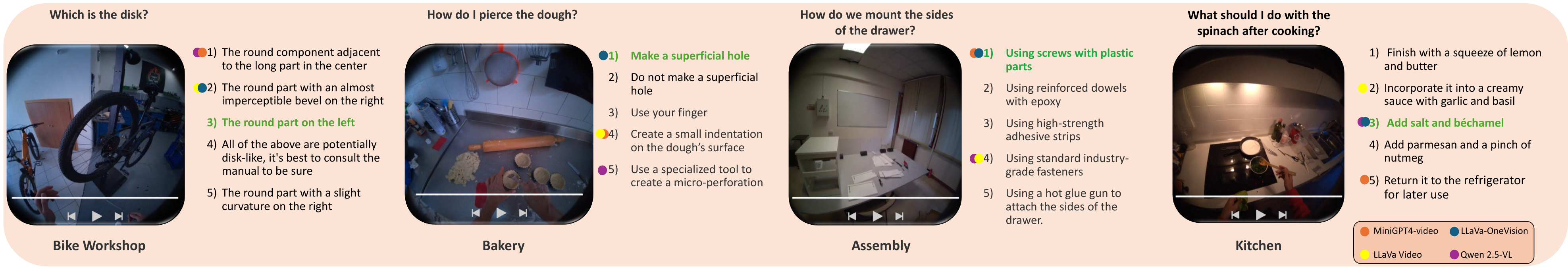}
    \caption{Qualitative results of the proposed VQA benchmark. Correct answer in \textcolor{Green}{green}, baselines predictions marked with colors.}
    \label{fig:qualitative}
    \vspace{-2mm}
\end{figure*}

Qualitative results are shown in Figures \ref{fig:qualitative_2}
and \ref{fig:qualitative}.


%% file: main.bbl
\begin{thebibliography}{73}
\providecommand{\natexlab}[1]{#1}
\providecommand{\url}[1]{\texttt{#1}}
\expandafter\ifx\csname urlstyle\endcsname\relax
  \providecommand{\doi}[1]{doi: #1}\else
  \providecommand{\doi}{doi: \begingroup \urlstyle{rm}\Url}\fi

\bibitem[ari()]{aria_mps}
Aria mps.
\newblock \url{https://facebookresearch.github.io/projectaria_tools/docs/ARK/mps}.

\bibitem[eHo()]{eHow}
ehow.
\newblock \url{https://www.ehow.com/}.

\bibitem[lla({\natexlab{a}})]{llama3.1_instruct}
Llama 3.1 instruct, {\natexlab{a}}.
\newblock \url{https://huggingface.co/meta-llama/Llama-3.1-8B-Instruct}.

\bibitem[lla({\natexlab{b}})]{llama3.3_instruct_turbo}
Llama 3.3 instruct turbo, {\natexlab{b}}.
\newblock \url{https://huggingface.co/meta-llama/Llama-3.3-70B-Instruct}.

\bibitem[wik()]{wikihow}
wikihow.
\newblock \url{https://www.wikihow.com/Main-Page}.

\bibitem[you()]{youtube}
youtube.
\newblock \url{https://www.youtube.com/}.

\bibitem[Ashutosh et~al.(2024)Ashutosh, Ramakrishnan, Afouras, and Grauman]{ashutosh2024video}
Kumar Ashutosh, Santhosh~Kumar Ramakrishnan, Triantafyllos Afouras, and Kristen Grauman.
\newblock Video-mined task graphs for keystep recognition in instructional videos.
\newblock \emph{Advances in Neural Information Processing Systems}, 36, 2024.

\bibitem[Ataallah et~al.(2024)Ataallah, Shen, Abdelrahman, Sleiman, Zhu, Ding, and Elhoseiny]{ataallah2024minigpt4videoadvancingmultimodalllms}
Kirolos Ataallah, Xiaoqian Shen, Eslam Abdelrahman, Essam Sleiman, Deyao Zhu, Jian Ding, and Mohamed Elhoseiny.
\newblock Minigpt4-video: Advancing multimodal llms for video understanding with interleaved visual-textual tokens, 2024.

\bibitem[Bai et~al.(2025)Bai, Chen, Liu, Wang, Ge, Song, Dang, Wang, Wang, Tang, Zhong, Zhu, Yang, Li, Wan, Wang, Ding, Fu, Xu, Ye, Zhang, Xie, Cheng, Zhang, Yang, Xu, and Lin]{Qwen2.5-VL}
Shuai Bai, Keqin Chen, Xuejing Liu, Jialin Wang, Wenbin Ge, Sibo Song, Kai Dang, Peng Wang, Shijie Wang, Jun Tang, Humen Zhong, Yuanzhi Zhu, Mingkun Yang, Zhaohai Li, Jianqiang Wan, Pengfei Wang, Wei Ding, Zheren Fu, Yiheng Xu, Jiabo Ye, Xi Zhang, Tianbao Xie, Zesen Cheng, Hang Zhang, Zhibo Yang, Haiyang Xu, and Junyang Lin.
\newblock Qwen2.5-vl technical report.
\newblock \emph{arXiv preprint arXiv:2502.13923}, 2025.

\bibitem[Bi et~al.(2024)Bi, Tang, Song, Vosoughi, Nguyen, and Xu]{Bi2024ARXIV}
Jing Bi, Yunlong Tang, Luchuan Song, Ali Vosoughi, Nguyen Nguyen, and Chenliang Xu.
\newblock {EAGLE:} egocentric {AGgregated} language-video engine.
\newblock \emph{arXiv}, 2409.17523, 2024.

\bibitem[Chen et~al.(2023)Chen, Ge, Ge, Ding, Li, Wang, Xu, Shan, and Liu]{chen2023egoplan}
Yi Chen, Yuying Ge, Yixiao Ge, Mingyu Ding, Bohao Li, Rui Wang, Ruifeng Xu, Ying Shan, and Xihui Liu.
\newblock Egoplan-bench: Benchmarking multimodal large language models for human-level planning.
\newblock \emph{arXiv preprint arXiv:2312.06722}, 2023.

\bibitem[Chen et~al.(2024)Chen, Ge, Ge, Ding, Li, Wang, Xu, Shan, and Liu]{Chen2024ARXIV}
Yi Chen, Yuying Ge, Yixiao Ge, Mingyu Ding, Bohao Li, Rui Wang, Ruifeng Xu, Ying Shan, and Xihui Liu.
\newblock {EgoPlan-Bench:} benchmarking multimodal large language models for human-level planning.
\newblock \emph{arXiv}, 2312.06722, 2024.

\bibitem[Cheng et~al.(2024{\natexlab{a}})Cheng, Fang, Yu, Zhou, Li, Tian, Li, Han, and Liu]{Cheng2024ARXIV}
Sijie Cheng, Kechen Fang, Yangyang Yu, Sicheng Zhou, Bohao Li, Ye Tian, Tingguang Li, Lei Han, and Yang Liu.
\newblock {VidEgoThink:} assessing egocentric video understanding capabilities for embodied {AI}.
\newblock \emph{arXiv}, 2410.11623, 2024{\natexlab{a}}.

\bibitem[Cheng et~al.(2024{\natexlab{b}})Cheng, Guo, Wu, Fang, Li, Liu, and Liu]{Cheng2024CVPR}
Sijie Cheng, Zhicheng Guo, Jingwen Wu, Kechen Fang, Peng Li, Huaping Liu, and Yang Liu.
\newblock {EgoThink:} evaluating first-person perspective thinking capability of vision-language models.
\newblock In \emph{Computer Vision and Pattern Recognition (CVPR)}, 2024{\natexlab{b}}.

\bibitem[Dai et~al.(2024)Dai, Li, Li, Tiong, Zhao, Wang, Li, Fung, and Hoi]{instruct_blip}
Wenliang Dai, Junnan Li, Dongxu Li, Anthony Meng~Huat Tiong, Junqi Zhao, Weisheng Wang, Boyang Li, Pascale Fung, and Steven Hoi.
\newblock Instructblip: towards general-purpose vision-language models with instruction tuning.
\newblock In \emph{Proceedings of the 37th International Conference on Neural Information Processing Systems}, Red Hook, NY, USA, 2024. Curran Associates Inc.

\bibitem[Damen et~al.(2018)Damen, Doughty, Farinella, Fidler, Furnari, Kazakos, Moltisanti, Munro, Perrett, Price, et~al.]{damen2018scaling}
Dima Damen, Hazel Doughty, Giovanni~Maria Farinella, Sanja Fidler, Antonino Furnari, Evangelos Kazakos, Davide Moltisanti, Jonathan Munro, Toby Perrett, Will Price, et~al.
\newblock Scaling egocentric vision: The epic-kitchens dataset.
\newblock In \emph{Proceedings of the European conference on computer vision (ECCV)}, pages 720--736, 2018.

\bibitem[Damen et~al.(2022)Damen, Doughty, Farinella, Furnari, Ma, Kazakos, Moltisanti, Munro, Perrett, Price, and Wray]{Damen2022RESCALING}
Dima Damen, Hazel Doughty, Giovanni~Maria Farinella, Antonino Furnari, Jian Ma, Evangelos Kazakos, Davide Moltisanti, Jonathan Munro, Toby Perrett, Will Price, and Michael Wray.
\newblock Rescaling egocentric vision: Collection, pipeline and challenges for epic-kitchens-100.
\newblock \emph{International Journal of Computer Vision (IJCV)}, 2022.

\bibitem[Datta et~al.(2022)Datta, Dharur, Cartillier, Desai, Khanna, Batra, and Parikh]{Datta_Dharur_Cartillier_Desai_Khanna_Batra_Parikh_2022}
Samyak Datta, Sameer Dharur, Vincent Cartillier, Ruta Desai, Mukul Khanna, Dhruv Batra, and Devi Parikh.
\newblock Episodic memory question answering.
\newblock 2022.

\bibitem[DeepSeek-AI et~al.(2025)DeepSeek-AI, Guo, Yang, Zhang, Song, Zhang, Xu, Zhu, Ma, Wang, Bi, Zhang, Yu, Wu, Wu, Gou, Shao, Li, Gao, Liu, Xue, Wang, Wu, Feng, Lu, Zhao, Deng, Zhang, Ruan, Dai, Chen, Ji, Li, Lin, Dai, Luo, Hao, Chen, Li, Zhang, Bao, Xu, Wang, Ding, Xin, Gao, Qu, Li, Guo, Li, Wang, Chen, Yuan, Qiu, Li, Cai, Ni, Liang, Chen, Dong, Hu, Gao, Guan, Huang, Yu, Wang, Zhang, Zhao, Wang, Zhang, Xu, Xia, Zhang, Zhang, Tang, Li, Wang, Li, Tian, Huang, Zhang, Wang, Chen, Du, Ge, Zhang, Pan, Wang, Chen, Jin, Chen, Lu, Zhou, Chen, Ye, Wang, Yu, Zhou, Pan, Li, Zhou, Wu, Ye, Yun, Pei, Sun, Wang, Zeng, Zhao, Liu, Liang, Gao, Yu, Zhang, Xiao, An, Liu, Wang, Chen, Nie, Cheng, Liu, Xie, Liu, Yang, Li, Su, Lin, Li, Jin, Shen, Chen, Sun, Wang, Song, Zhou, Wang, Shan, Li, Wang, Wei, Zhang, Xu, Li, Zhao, Sun, Wang, Yu, Zhang, Shi, Xiong, He, Piao, Wang, Tan, Ma, Liu, Guo, Ou, Wang, Gong, Zou, He, Xiong, Luo, You, Liu, Zhou, Zhu, Xu, Huang, Li, Zheng, Zhu, Ma, Tang, Zha, Yan, Ren, Ren, Sha, Fu, Xu, Xie, Zhang,
  Hao, Ma, Yan, Wu, Gu, Zhu, Liu, Li, Xie, Song, Pan, Huang, Xu, Zhang, and Zhang]{deepseekai2025deepseekr1}
DeepSeek-AI, Daya Guo, Dejian Yang, Haowei Zhang, Junxiao Song, Ruoyu Zhang, Runxin Xu, Qihao Zhu, Shirong Ma, Peiyi Wang, Xiao Bi, Xiaokang Zhang, Xingkai Yu, Yu Wu, Z.~F. Wu, Zhibin Gou, Zhihong Shao, Zhuoshu Li, Ziyi Gao, Aixin Liu, Bing Xue, Bingxuan Wang, Bochao Wu, Bei Feng, Chengda Lu, Chenggang Zhao, Chengqi Deng, Chenyu Zhang, Chong Ruan, Damai Dai, Deli Chen, Dongjie Ji, Erhang Li, Fangyun Lin, Fucong Dai, Fuli Luo, Guangbo Hao, Guanting Chen, Guowei Li, H. Zhang, Han Bao, Hanwei Xu, Haocheng Wang, Honghui Ding, Huajian Xin, Huazuo Gao, Hui Qu, Hui Li, Jianzhong Guo, Jiashi Li, Jiawei Wang, Jingchang Chen, Jingyang Yuan, Junjie Qiu, Junlong Li, J.~L. Cai, Jiaqi Ni, Jian Liang, Jin Chen, Kai Dong, Kai Hu, Kaige Gao, Kang Guan, Kexin Huang, Kuai Yu, Lean Wang, Lecong Zhang, Liang Zhao, Litong Wang, Liyue Zhang, Lei Xu, Leyi Xia, Mingchuan Zhang, Minghua Zhang, Minghui Tang, Meng Li, Miaojun Wang, Mingming Li, Ning Tian, Panpan Huang, Peng Zhang, Qiancheng Wang, Qinyu Chen, Qiushi Du, Ruiqi Ge, Ruisong
  Zhang, Ruizhe Pan, Runji Wang, R.~J. Chen, R.~L. Jin, Ruyi Chen, Shanghao Lu, Shangyan Zhou, Shanhuang Chen, Shengfeng Ye, Shiyu Wang, Shuiping Yu, Shunfeng Zhou, Shuting Pan, S.~S. Li, Shuang Zhou, Shaoqing Wu, Shengfeng Ye, Tao Yun, Tian Pei, Tianyu Sun, T. Wang, Wangding Zeng, Wanjia Zhao, Wen Liu, Wenfeng Liang, Wenjun Gao, Wenqin Yu, Wentao Zhang, W.~L. Xiao, Wei An, Xiaodong Liu, Xiaohan Wang, Xiaokang Chen, Xiaotao Nie, Xin Cheng, Xin Liu, Xin Xie, Xingchao Liu, Xinyu Yang, Xinyuan Li, Xuecheng Su, Xuheng Lin, X.~Q. Li, Xiangyue Jin, Xiaojin Shen, Xiaosha Chen, Xiaowen Sun, Xiaoxiang Wang, Xinnan Song, Xinyi Zhou, Xianzu Wang, Xinxia Shan, Y.~K. Li, Y.~Q. Wang, Y.~X. Wei, Yang Zhang, Yanhong Xu, Yao Li, Yao Zhao, Yaofeng Sun, Yaohui Wang, Yi Yu, Yichao Zhang, Yifan Shi, Yiliang Xiong, Ying He, Yishi Piao, Yisong Wang, Yixuan Tan, Yiyang Ma, Yiyuan Liu, Yongqiang Guo, Yuan Ou, Yuduan Wang, Yue Gong, Yuheng Zou, Yujia He, Yunfan Xiong, Yuxiang Luo, Yuxiang You, Yuxuan Liu, Yuyang Zhou, Y.~X. Zhu,
  Yanhong Xu, Yanping Huang, Yaohui Li, Yi Zheng, Yuchen Zhu, Yunxian Ma, Ying Tang, Yukun Zha, Yuting Yan, Z.~Z. Ren, Zehui Ren, Zhangli Sha, Zhe Fu, Zhean Xu, Zhenda Xie, Zhengyan Zhang, Zhewen Hao, Zhicheng Ma, Zhigang Yan, Zhiyu Wu, Zihui Gu, Zijia Zhu, Zijun Liu, Zilin Li, Ziwei Xie, Ziyang Song, Zizheng Pan, Zhen Huang, Zhipeng Xu, Zhongyu Zhang, and Zhen Zhang.
\newblock Deepseek-r1: Incentivizing reasoning capability in llms via reinforcement learning, 2025.

\bibitem[Dong et~al.(2024)Dong, Beedu, Sheinkopf, and Essa]{Dong2024ARXIV}
Zhikang Dong, Apoorva Beedu, Jason Sheinkopf, and Irfan Essa.
\newblock Mamba fusion: Learning actions through questioning.
\newblock \emph{arXiv}, 2409.11513, 2024.

\bibitem[Doughty et~al.(2018)Doughty, Damen, and Mayol-Cuevas]{doughty2018s}
Hazel Doughty, Dima Damen, and Walterio Mayol-Cuevas.
\newblock Who's better? who's best? pairwise deep ranking for skill determination.
\newblock In \emph{Proceedings of the IEEE conference on computer vision and pattern recognition}, pages 6057--6066, 2018.

\bibitem[Doughty et~al.(2019)Doughty, Mayol-Cuevas, and Damen]{Doughty_2019_CVPR}
Hazel Doughty, Walterio Mayol-Cuevas, and Dima Damen.
\newblock The pros and cons: Rank-aware temporal attention for skill determination in long videos.
\newblock In \emph{Proceedings of the IEEE/CVF Conference on Computer Vision and Pattern Recognition (CVPR)}, 2019.

\bibitem[Dubey et~al.(2024)Dubey, Jauhri, Pandey, Kadian, Al-Dahle, Letman, Mathur, Schelten, Yang, Fan, Goyal, Hartshorn, Yang, Mitra, Sravankumar, Korenev, Hinsvark, Rao, Zhang, Rodriguez, Gregerson, Spataru, Roziere, Biron, Tang, Chern, Caucheteux, Nayak, Bi, Marra, McConnell, Keller, Touret, Wu, Wong, Ferrer, Nikolaidis, Allonsius, Song, Pintz, Livshits, Esiobu, Choudhary, Mahajan, Garcia-Olano, Perino, Hupkes, Lakomkin, AlBadawy, Lobanova, Dinan, Smith, Radenovic, Zhang, Synnaeve, Lee, Anderson, Nail, Mialon, Pang, Cucurell, Nguyen, Korevaar, Xu, Touvron, Zarov, Ibarra, Kloumann, Misra, Evtimov, Copet, Lee, Geffert, Vranes, Park, Mahadeokar, Shah, van~der Linde, Billock, Hong, Lee, Fu, Chi, Huang, Liu, Wang, Yu, Bitton, Spisak, Park, Rocca, Johnstun, Saxe, Jia, Alwala, Upasani, Plawiak, Li, Heafield, Stone, El-Arini, Iyer, Malik, Chiu, Bhalla, Rantala-Yeary, van~der Maaten, Chen, Tan, Jenkins, Martin, Madaan, Malo, Blecher, Landzaat, de~Oliveira, Muzzi, Pasupuleti, Singh, Paluri, Kardas, Oldham, Rita,
  Pavlova, Kambadur, Lewis, Si, Singh, Hassan, Goyal, Torabi, Bashlykov, Bogoychev, Chatterji, Duchenne, Çelebi, Alrassy, Zhang, Li, Vasic, Weng, Bhargava, Dubal, Krishnan, Koura, Xu, He, Dong, Srinivasan, Ganapathy, Calderer, Cabral, Stojnic, Raileanu, Girdhar, Patel, Sauvestre, Polidoro, Sumbaly, Taylor, Silva, Hou, Wang, Hosseini, Chennabasappa, Singh, Bell, Kim, Edunov, Nie, Narang, Raparthy, Shen, Wan, Bhosale, Zhang, Vandenhende, Batra, Whitman, Sootla, Collot, Gururangan, Borodinsky, Herman, Fowler, Sheasha, Georgiou, Scialom, Speckbacher, Mihaylov, Xiao, Karn, Goswami, Gupta, Ramanathan, Kerkez, Gonguet, Do, Vogeti, Petrovic, Chu, Xiong, Fu, Meers, Martinet, Wang, Tan, Xie, Jia, Wang, Goldschlag, Gaur, Babaei, Wen, Song, Zhang, Li, Mao, Coudert, Yan, Chen, Papakipos, Singh, Grattafiori, Jain, Kelsey, Shajnfeld, Gangidi, Victoria, Goldstand, Menon, Sharma, Boesenberg, Vaughan, Baevski, Feinstein, Kallet, Sangani, Yunus, Lupu, Alvarado, Caples, Gu, Ho, Poulton, Ryan, Ramchandani, Franco, Saraf,
  Chowdhury, Gabriel, Bharambe, Eisenman, Yazdan, James, Maurer, Leonhardi, Huang, Loyd, Paola, Paranjape, Liu, Wu, Ni, Hancock, Wasti, Spence, Stojkovic, Gamido, Montalvo, Parker, Burton, Mejia, Wang, Kim, Zhou, Hu, Chu, Cai, Tindal, Feichtenhofer, Civin, Beaty, Kreymer, Li, Wyatt, Adkins, Xu, Testuggine, David, Parikh, Liskovich, Foss, Wang, Le, Holland, Dowling, Jamil, Montgomery, Presani, Hahn, Wood, Brinkman, Arcaute, Dunbar, Smothers, Sun, Kreuk, Tian, Ozgenel, Caggioni, Guzmán, Kanayet, Seide, Florez, Schwarz, Badeer, Swee, Halpern, Thattai, Herman, Sizov, Guangyi, Zhang, Lakshminarayanan, Shojanazeri, Zou, Wang, Zha, Habeeb, Rudolph, Suk, Aspegren, Goldman, Damlaj, Molybog, Tufanov, Veliche, Gat, Weissman, Geboski, Kohli, Asher, Gaya, Marcus, Tang, Chan, Zhen, Reizenstein, Teboul, Zhong, Jin, Yang, Cummings, Carvill, Shepard, McPhie, Torres, Ginsburg, Wang, Wu, U, Saxena, Prasad, Khandelwal, Zand, Matosich, Veeraraghavan, Michelena, Li, Huang, Chawla, Lakhotia, Huang, Chen, Garg, A, Silva, Bell,
  Zhang, Guo, Yu, Moshkovich, Wehrstedt, Khabsa, Avalani, Bhatt, Tsimpoukelli, Mankus, Hasson, Lennie, Reso, Groshev, Naumov, Lathi, Keneally, Seltzer, Valko, Restrepo, Patel, Vyatskov, Samvelyan, Clark, Macey, Wang, Hermoso, Metanat, Rastegari, Bansal, Santhanam, Parks, White, Bawa, Singhal, Egebo, Usunier, Laptev, Dong, Zhang, Cheng, Chernoguz, Hart, Salpekar, Kalinli, Kent, Parekh, Saab, Balaji, Rittner, Bontrager, Roux, Dollar, Zvyagina, Ratanchandani, Yuvraj, Liang, Alao, Rodriguez, Ayub, Murthy, Nayani, Mitra, Li, Hogan, Battey, Wang, Maheswari, Howes, Rinott, Bondu, Datta, Chugh, Hunt, Dhillon, Sidorov, Pan, Verma, Yamamoto, Ramaswamy, Lindsay, Lindsay, Feng, Lin, Zha, Shankar, Zhang, Zhang, Wang, Agarwal, Sajuyigbe, Chintala, Max, Chen, Kehoe, Satterfield, Govindaprasad, Gupta, Cho, Virk, Subramanian, Choudhury, Goldman, Remez, Glaser, Best, Kohler, Robinson, Li, Zhang, Matthews, Chou, Shaked, Vontimitta, Ajayi, Montanez, Mohan, Kumar, Mangla, Albiero, Ionescu, Poenaru, Mihailescu, Ivanov, Li, Wang,
  Jiang, Bouaziz, Constable, Tang, Wang, Wu, Wang, Xia, Wu, Gao, Chen, Hu, Jia, Qi, Li, Zhang, Zhang, Adi, Nam, Yu, Wang, Hao, Qian, He, Rait, DeVito, Rosnbrick, Wen, Yang, and Zhao]{dubey2024llama3herdmodels}
Abhimanyu Dubey, Abhinav Jauhri, Abhinav Pandey, Abhishek Kadian, Ahmad Al-Dahle, Aiesha Letman, Akhil Mathur, Alan Schelten, Amy Yang, Angela Fan, Anirudh Goyal, Anthony Hartshorn, Aobo Yang, Archi Mitra, Archie Sravankumar, Artem Korenev, Arthur Hinsvark, Arun Rao, Aston Zhang, Aurelien Rodriguez, Austen Gregerson, Ava Spataru, Baptiste Roziere, Bethany Biron, Binh Tang, Bobbie Chern, Charlotte Caucheteux, Chaya Nayak, Chloe Bi, Chris Marra, Chris McConnell, Christian Keller, Christophe Touret, Chunyang Wu, Corinne Wong, Cristian~Canton Ferrer, Cyrus Nikolaidis, Damien Allonsius, Daniel Song, Danielle Pintz, Danny Livshits, David Esiobu, Dhruv Choudhary, Dhruv Mahajan, Diego Garcia-Olano, Diego Perino, Dieuwke Hupkes, Egor Lakomkin, Ehab AlBadawy, Elina Lobanova, Emily Dinan, Eric~Michael Smith, Filip Radenovic, Frank Zhang, Gabriel Synnaeve, Gabrielle Lee, Georgia~Lewis Anderson, Graeme Nail, Gregoire Mialon, Guan Pang, Guillem Cucurell, Hailey Nguyen, Hannah Korevaar, Hu Xu, Hugo Touvron, Iliyan Zarov,
  Imanol~Arrieta Ibarra, Isabel Kloumann, Ishan Misra, Ivan Evtimov, Jade Copet, Jaewon Lee, Jan Geffert, Jana Vranes, Jason Park, Jay Mahadeokar, Jeet Shah, Jelmer van~der Linde, Jennifer Billock, Jenny Hong, Jenya Lee, Jeremy Fu, Jianfeng Chi, Jianyu Huang, Jiawen Liu, Jie Wang, Jiecao Yu, Joanna Bitton, Joe Spisak, Jongsoo Park, Joseph Rocca, Joshua Johnstun, Joshua Saxe, Junteng Jia, Kalyan~Vasuden Alwala, Kartikeya Upasani, Kate Plawiak, Ke Li, Kenneth Heafield, Kevin Stone, Khalid El-Arini, Krithika Iyer, Kshitiz Malik, Kuenley Chiu, Kunal Bhalla, Lauren Rantala-Yeary, Laurens van~der Maaten, Lawrence Chen, Liang Tan, Liz Jenkins, Louis Martin, Lovish Madaan, Lubo Malo, Lukas Blecher, Lukas Landzaat, Luke de Oliveira, Madeline Muzzi, Mahesh Pasupuleti, Mannat Singh, Manohar Paluri, Marcin Kardas, Mathew Oldham, Mathieu Rita, Maya Pavlova, Melanie Kambadur, Mike Lewis, Min Si, Mitesh~Kumar Singh, Mona Hassan, Naman Goyal, Narjes Torabi, Nikolay Bashlykov, Nikolay Bogoychev, Niladri Chatterji, Olivier
  Duchenne, Onur Çelebi, Patrick Alrassy, Pengchuan Zhang, Pengwei Li, Petar Vasic, Peter Weng, Prajjwal Bhargava, Pratik Dubal, Praveen Krishnan, Punit~Singh Koura, Puxin Xu, Qing He, Qingxiao Dong, Ragavan Srinivasan, Raj Ganapathy, Ramon Calderer, Ricardo~Silveira Cabral, Robert Stojnic, Roberta Raileanu, Rohit Girdhar, Rohit Patel, Romain Sauvestre, Ronnie Polidoro, Roshan Sumbaly, Ross Taylor, Ruan Silva, Rui Hou, Rui Wang, Saghar Hosseini, Sahana Chennabasappa, Sanjay Singh, Sean Bell, Seohyun~Sonia Kim, Sergey Edunov, Shaoliang Nie, Sharan Narang, Sharath Raparthy, Sheng Shen, Shengye Wan, Shruti Bhosale, Shun Zhang, Simon Vandenhende, Soumya Batra, Spencer Whitman, Sten Sootla, Stephane Collot, Suchin Gururangan, Sydney Borodinsky, Tamar Herman, Tara Fowler, Tarek Sheasha, Thomas Georgiou, Thomas Scialom, Tobias Speckbacher, Todor Mihaylov, Tong Xiao, Ujjwal Karn, Vedanuj Goswami, Vibhor Gupta, Vignesh Ramanathan, Viktor Kerkez, Vincent Gonguet, Virginie Do, Vish Vogeti, Vladan Petrovic, Weiwei Chu,
  Wenhan Xiong, Wenyin Fu, Whitney Meers, Xavier Martinet, Xiaodong Wang, Xiaoqing~Ellen Tan, Xinfeng Xie, Xuchao Jia, Xuewei Wang, Yaelle Goldschlag, Yashesh Gaur, Yasmine Babaei, Yi Wen, Yiwen Song, Yuchen Zhang, Yue Li, Yuning Mao, Zacharie~Delpierre Coudert, Zheng Yan, Zhengxing Chen, Zoe Papakipos, Aaditya Singh, Aaron Grattafiori, Abha Jain, Adam Kelsey, Adam Shajnfeld, Adithya Gangidi, Adolfo Victoria, Ahuva Goldstand, Ajay Menon, Ajay Sharma, Alex Boesenberg, Alex Vaughan, Alexei Baevski, Allie Feinstein, Amanda Kallet, Amit Sangani, Anam Yunus, Andrei Lupu, Andres Alvarado, Andrew Caples, Andrew Gu, Andrew Ho, Andrew Poulton, Andrew Ryan, Ankit Ramchandani, Annie Franco, Aparajita Saraf, Arkabandhu Chowdhury, Ashley Gabriel, Ashwin Bharambe, Assaf Eisenman, Azadeh Yazdan, Beau James, Ben Maurer, Benjamin Leonhardi, Bernie Huang, Beth Loyd, Beto~De Paola, Bhargavi Paranjape, Bing Liu, Bo Wu, Boyu Ni, Braden Hancock, Bram Wasti, Brandon Spence, Brani Stojkovic, Brian Gamido, Britt Montalvo, Carl
  Parker, Carly Burton, Catalina Mejia, Changhan Wang, Changkyu Kim, Chao Zhou, Chester Hu, Ching-Hsiang Chu, Chris Cai, Chris Tindal, Christoph Feichtenhofer, Damon Civin, Dana Beaty, Daniel Kreymer, Daniel Li, Danny Wyatt, David Adkins, David Xu, Davide Testuggine, Delia David, Devi Parikh, Diana Liskovich, Didem Foss, Dingkang Wang, Duc Le, Dustin Holland, Edward Dowling, Eissa Jamil, Elaine Montgomery, Eleonora Presani, Emily Hahn, Emily Wood, Erik Brinkman, Esteban Arcaute, Evan Dunbar, Evan Smothers, Fei Sun, Felix Kreuk, Feng Tian, Firat Ozgenel, Francesco Caggioni, Francisco Guzmán, Frank Kanayet, Frank Seide, Gabriela~Medina Florez, Gabriella Schwarz, Gada Badeer, Georgia Swee, Gil Halpern, Govind Thattai, Grant Herman, Grigory Sizov, Guangyi, Zhang, Guna Lakshminarayanan, Hamid Shojanazeri, Han Zou, Hannah Wang, Hanwen Zha, Haroun Habeeb, Harrison Rudolph, Helen Suk, Henry Aspegren, Hunter Goldman, Ibrahim Damlaj, Igor Molybog, Igor Tufanov, Irina-Elena Veliche, Itai Gat, Jake Weissman, James
  Geboski, James Kohli, Japhet Asher, Jean-Baptiste Gaya, Jeff Marcus, Jeff Tang, Jennifer Chan, Jenny Zhen, Jeremy Reizenstein, Jeremy Teboul, Jessica Zhong, Jian Jin, Jingyi Yang, Joe Cummings, Jon Carvill, Jon Shepard, Jonathan McPhie, Jonathan Torres, Josh Ginsburg, Junjie Wang, Kai Wu, Kam~Hou U, Karan Saxena, Karthik Prasad, Kartikay Khandelwal, Katayoun Zand, Kathy Matosich, Kaushik Veeraraghavan, Kelly Michelena, Keqian Li, Kun Huang, Kunal Chawla, Kushal Lakhotia, Kyle Huang, Lailin Chen, Lakshya Garg, Lavender A, Leandro Silva, Lee Bell, Lei Zhang, Liangpeng Guo, Licheng Yu, Liron Moshkovich, Luca Wehrstedt, Madian Khabsa, Manav Avalani, Manish Bhatt, Maria Tsimpoukelli, Martynas Mankus, Matan Hasson, Matthew Lennie, Matthias Reso, Maxim Groshev, Maxim Naumov, Maya Lathi, Meghan Keneally, Michael~L. Seltzer, Michal Valko, Michelle Restrepo, Mihir Patel, Mik Vyatskov, Mikayel Samvelyan, Mike Clark, Mike Macey, Mike Wang, Miquel~Jubert Hermoso, Mo Metanat, Mohammad Rastegari, Munish Bansal, Nandhini
  Santhanam, Natascha Parks, Natasha White, Navyata Bawa, Nayan Singhal, Nick Egebo, Nicolas Usunier, Nikolay~Pavlovich Laptev, Ning Dong, Ning Zhang, Norman Cheng, Oleg Chernoguz, Olivia Hart, Omkar Salpekar, Ozlem Kalinli, Parkin Kent, Parth Parekh, Paul Saab, Pavan Balaji, Pedro Rittner, Philip Bontrager, Pierre Roux, Piotr Dollar, Polina Zvyagina, Prashant Ratanchandani, Pritish Yuvraj, Qian Liang, Rachad Alao, Rachel Rodriguez, Rafi Ayub, Raghotham Murthy, Raghu Nayani, Rahul Mitra, Raymond Li, Rebekkah Hogan, Robin Battey, Rocky Wang, Rohan Maheswari, Russ Howes, Ruty Rinott, Sai~Jayesh Bondu, Samyak Datta, Sara Chugh, Sara Hunt, Sargun Dhillon, Sasha Sidorov, Satadru Pan, Saurabh Verma, Seiji Yamamoto, Sharadh Ramaswamy, Shaun Lindsay, Shaun Lindsay, Sheng Feng, Shenghao Lin, Shengxin~Cindy Zha, Shiva Shankar, Shuqiang Zhang, Shuqiang Zhang, Sinong Wang, Sneha Agarwal, Soji Sajuyigbe, Soumith Chintala, Stephanie Max, Stephen Chen, Steve Kehoe, Steve Satterfield, Sudarshan Govindaprasad, Sumit Gupta,
  Sungmin Cho, Sunny Virk, Suraj Subramanian, Sy Choudhury, Sydney Goldman, Tal Remez, Tamar Glaser, Tamara Best, Thilo Kohler, Thomas Robinson, Tianhe Li, Tianjun Zhang, Tim Matthews, Timothy Chou, Tzook Shaked, Varun Vontimitta, Victoria Ajayi, Victoria Montanez, Vijai Mohan, Vinay~Satish Kumar, Vishal Mangla, Vítor Albiero, Vlad Ionescu, Vlad Poenaru, Vlad~Tiberiu Mihailescu, Vladimir Ivanov, Wei Li, Wenchen Wang, Wenwen Jiang, Wes Bouaziz, Will Constable, Xiaocheng Tang, Xiaofang Wang, Xiaojian Wu, Xiaolan Wang, Xide Xia, Xilun Wu, Xinbo Gao, Yanjun Chen, Ye Hu, Ye Jia, Ye Qi, Yenda Li, Yilin Zhang, Ying Zhang, Yossi Adi, Youngjin Nam, Yu, Wang, Yuchen Hao, Yundi Qian, Yuzi He, Zach Rait, Zachary DeVito, Zef Rosnbrick, Zhaoduo Wen, Zhenyu Yang, and Zhiwei Zhao.
\newblock The llama 3 herd of models, 2024.

\bibitem[Fan(2019)]{Fan_2019}
Chenyou Fan.
\newblock Egovqa - an egocentric video question answering benchmark dataset.
\newblock In \emph{2019 IEEE/CVF International Conference on Computer Vision Workshop (ICCVW)}, page 4359–4366, Seoul, Korea (South), 2019. IEEE.

\bibitem[Flaborea et~al.(2024)Flaborea, di~Melendugno, Plini, Scofano, De~Matteis, Furnari, Farinella, and Galasso]{Flaborea_2024_CVPR}
Alessandro Flaborea, Guido Maria~D'Amely di Melendugno, Leonardo Plini, Luca Scofano, Edoardo De~Matteis, Antonino Furnari, Giovanni~Maria Farinella, and Fabio Galasso.
\newblock Prego: Online mistake detection in procedural egocentric videos.
\newblock In \emph{Proceedings of the IEEE/CVF Conference on Computer Vision and Pattern Recognition (CVPR)}, pages 18483--18492, 2024.

\bibitem[Gao et~al.(2021)Gao, Wang, Bai, and Chen]{Gao_Wang_Bai_Chen_2021}
Difei Gao, Ruiping Wang, Ziyi Bai, and Xilin Chen.
\newblock Env-qa: A video question answering benchmark for comprehensive understanding of dynamic environments.
\newblock In \emph{2021 IEEE/CVF International Conference on Computer Vision (ICCV)}, page 1655–1665, Montreal, QC, Canada, 2021. IEEE.

\bibitem[Girdhar and Grauman(2021)]{girdhar2021anticipative}
Rohit Girdhar and Kristen Grauman.
\newblock Anticipative video transformer.
\newblock In \emph{Proceedings of the IEEE/CVF international conference on computer vision}, pages 13505--13515, 2021.

\bibitem[Grauman et~al.(2022)Grauman, Westbury, Byrne, Chavis, Furnari, Girdhar, Hamburger, Jiang, Liu, Liu, Martin, Nagarajan, Radosavovic, Ramakrishnan, Ryan, Sharma, Wray, Xu, Xu, Zhao, Bansal, Batra, Cartillier, Crane, Do, Doulaty, Erapalli, Feichtenhofer, Fragomeni, Fu, Gebreselasie, Gonz\'alez, Hillis, Huang, Huang, Jia, Khoo, Kol\'a\v{r}, Kottur, Kumar, Landini, Li, Li, Li, Mangalam, Modhugu, Munro, Murrell, Nishiyasu, Price, Ruiz, Ramazanova, Sari, Somasundaram, Southerland, Sugano, Tao, Vo, Wang, Wu, Yagi, Zhao, Zhu, Arbel\'aez, Crandall, Damen, Farinella, Fuegen, Ghanem, Ithapu, Jawahar, Joo, Kitani, Li, Newcombe, Oliva, Park, Rehg, Sato, Shi, Shou, Torralba, Torresani, Yan, and Malik]{Grauman_2022_CVPR}
Kristen Grauman, Andrew Westbury, Eugene Byrne, Zachary Chavis, Antonino Furnari, Rohit Girdhar, Jackson Hamburger, Hao Jiang, Miao Liu, Xingyu Liu, Miguel Martin, Tushar Nagarajan, Ilija Radosavovic, Santhosh~Kumar Ramakrishnan, Fiona Ryan, Jayant Sharma, Michael Wray, Mengmeng Xu, Eric~Zhongcong Xu, Chen Zhao, Siddhant Bansal, Dhruv Batra, Vincent Cartillier, Sean Crane, Tien Do, Morrie Doulaty, Akshay Erapalli, Christoph Feichtenhofer, Adriano Fragomeni, Qichen Fu, Abrham Gebreselasie, Cristina Gonz\'alez, James Hillis, Xuhua Huang, Yifei Huang, Wenqi Jia, Weslie Khoo, J\'achym Kol\'a\v{r}, Satwik Kottur, Anurag Kumar, Federico Landini, Chao Li, Yanghao Li, Zhenqiang Li, Karttikeya Mangalam, Raghava Modhugu, Jonathan Munro, Tullie Murrell, Takumi Nishiyasu, Will Price, Paola Ruiz, Merey Ramazanova, Leda Sari, Kiran Somasundaram, Audrey Southerland, Yusuke Sugano, Ruijie Tao, Minh Vo, Yuchen Wang, Xindi Wu, Takuma Yagi, Ziwei Zhao, Yunyi Zhu, Pablo Arbel\'aez, David Crandall, Dima Damen, Giovanni~Maria
  Farinella, Christian Fuegen, Bernard Ghanem, Vamsi~Krishna Ithapu, C.~V. Jawahar, Hanbyul Joo, Kris Kitani, Haizhou Li, Richard Newcombe, Aude Oliva, Hyun~Soo Park, James~M. Rehg, Yoichi Sato, Jianbo Shi, Mike~Zheng Shou, Antonio Torralba, Lorenzo Torresani, Mingfei Yan, and Jitendra Malik.
\newblock Ego4d: Around the world in 3,000 hours of egocentric video.
\newblock In \emph{Proceedings of the IEEE/CVF Conference on Computer Vision and Pattern Recognition (CVPR)}, pages 18995--19012, 2022.

\bibitem[Grauman et~al.(2024)Grauman, Westbury, Torresani, Kitani, Malik, Afouras, Ashutosh, Baiyya, Bansal, Boote, Byrne, Chavis, Chen, Cheng, Chu, Crane, Dasgupta, Dong, Escobar, Forigua, Gebreselasie, Haresh, Huang, Islam, Jain, Khirodkar, Kukreja, Liang, Liu, Majumder, Mao, Martin, Mavroudi, Nagarajan, Ragusa, Ramakrishnan, Seminara, Somayazulu, Song, Su, Xue, Zhang, Zhang, Castillo, Chen, Fu, Furuta, Gonzalez, Gupta, Hu, Huang, Huang, Khoo, Kumar, Kuo, Lakhavani, Liu, Luo, Luo, Meredith, Miller, Oguntola, Pan, Peng, Pramanick, Ramazanova, Ryan, Shan, Somasundaram, Song, Southerland, Tateno, Wang, Wang, Yagi, Yan, Yang, Yu, Zha, Zhao, Zhao, Zhu, Zhuo, Arbelaez, Bertasius, Damen, Engel, Farinella, Furnari, Ghanem, Hoffman, Jawahar, Newcombe, Park, Rehg, Sato, Savva, Shi, Shou, and Wray]{Grauman_2024_CVPR}
Kristen Grauman, Andrew Westbury, Lorenzo Torresani, Kris Kitani, Jitendra Malik, Triantafyllos Afouras, Kumar Ashutosh, Vijay Baiyya, Siddhant Bansal, Bikram Boote, Eugene Byrne, Zach Chavis, Joya Chen, Feng Cheng, Fu-Jen Chu, Sean Crane, Avijit Dasgupta, Jing Dong, Maria Escobar, Cristhian Forigua, Abrham Gebreselasie, Sanjay Haresh, Jing Huang, Md~Mohaiminul Islam, Suyog Jain, Rawal Khirodkar, Devansh Kukreja, Kevin~J Liang, Jia-Wei Liu, Sagnik Majumder, Yongsen Mao, Miguel Martin, Effrosyni Mavroudi, Tushar Nagarajan, Francesco Ragusa, Santhosh~Kumar Ramakrishnan, Luigi Seminara, Arjun Somayazulu, Yale Song, Shan Su, Zihui Xue, Edward Zhang, Jinxu Zhang, Angela Castillo, Changan Chen, Xinzhu Fu, Ryosuke Furuta, Cristina Gonzalez, Prince Gupta, Jiabo Hu, Yifei Huang, Yiming Huang, Weslie Khoo, Anush Kumar, Robert Kuo, Sach Lakhavani, Miao Liu, Mi Luo, Zhengyi Luo, Brighid Meredith, Austin Miller, Oluwatumininu Oguntola, Xiaqing Pan, Penny Peng, Shraman Pramanick, Merey Ramazanova, Fiona Ryan, Wei Shan,
  Kiran Somasundaram, Chenan Song, Audrey Southerland, Masatoshi Tateno, Huiyu Wang, Yuchen Wang, Takuma Yagi, Mingfei Yan, Xitong Yang, Zecheng Yu, Shengxin~Cindy Zha, Chen Zhao, Ziwei Zhao, Zhifan Zhu, Jeff Zhuo, Pablo Arbelaez, Gedas Bertasius, Dima Damen, Jakob Engel, Giovanni~Maria Farinella, Antonino Furnari, Bernard Ghanem, Judy Hoffman, C.V. Jawahar, Richard Newcombe, Hyun~Soo Park, James~M. Rehg, Yoichi Sato, Manolis Savva, Jianbo Shi, Mike~Zheng Shou, and Michael Wray.
\newblock Ego-exo4d: Understanding skilled human activity from first- and third-person perspectives.
\newblock In \emph{Proceedings of the IEEE/CVF Conference on Computer Vision and Pattern Recognition (CVPR)}, pages 19383--19400, 2024.

\bibitem[Hasegawa et~al.(2024{\natexlab{a}})Hasegawa, Imrattanatrai, Cheng, Asada, Holm, Wang, Fukuda, and Mitamura]{Hasegawa2024ARXIV}
Kimihiro Hasegawa, Wiradee Imrattanatrai, Zhi-Qi Cheng, Masaki Asada, Susan Holm, Yuran Wang, Ken Fukuda, and Teruko Mitamura.
\newblock {ProMQA:} question answering dataset for multimodal procedural activity understanding.
\newblock \emph{arXiv}, 2410.22211, 2024{\natexlab{a}}.

\bibitem[Hasegawa et~al.(2024{\natexlab{b}})Hasegawa, Imrattanatrai, Cheng, Asada, Holm, Wang, Fukuda, and Mitamura]{hasegawa2024promqaquestionansweringdataset}
Kimihiro Hasegawa, Wiradee Imrattanatrai, Zhi-Qi Cheng, Masaki Asada, Susan Holm, Yuran Wang, Ken Fukuda, and Teruko Mitamura.
\newblock Promqa: Question answering dataset for multimodal procedural activity understanding, 2024{\natexlab{b}}.

\bibitem[Islam et~al.(2024)Islam, Nagarajan, Wang, Chu, Kitani, Bertasius, and Yang]{islam2024propose}
Md~Mohaiminul Islam, Tushar Nagarajan, Huiyu Wang, Fu-Jen Chu, Kris Kitani, Gedas Bertasius, and Xitong Yang.
\newblock Propose, assess, search: Harnessing llms for goal-oriented planning in instructional videos.
\newblock In \emph{European Conference on Computer Vision}, 2024.

\bibitem[Jang et~al.(2019)Jang, Sullivan, Ludwig, Gilchrist, Damen, and Mayol-Cuevas]{jang2019epic}
Youngkyoon Jang, Brian Sullivan, Casimir Ludwig, Iain Gilchrist, Dima Damen, and Walterio Mayol-Cuevas.
\newblock Epic-tent: An egocentric video dataset for camping tent assembly.
\newblock In \emph{Proceedings of the IEEE/CVF International Conference on Computer Vision Workshops}, pages 0--0, 2019.

\bibitem[Jia et~al.(2020)Jia, Chen, Huang, Zhu, and Zhu]{lemma_dataset}
Baoxiong Jia, Yixin Chen, Siyuan Huang, Yixin Zhu, and Song-Chun Zhu.
\newblock Lemma: A multi-view dataset for learning multi-agent multi-task activities.
\newblock page 767–786, Berlin, Heidelberg, 2020. Springer-Verlag.

\bibitem[Jia et~al.(2022)Jia, Lei, Zhu, and Huang]{Jia_Lei_Zhu_Huang_2022}
Baoxiong Jia, Ting Lei, Song-Chun Zhu, and Siyuan Huang.
\newblock Egotaskqa: Understanding human tasks in egocentric videos.
\newblock \penalty0 (arXiv:2210.03929), 2022.
\newblock arXiv:2210.03929 [cs].

\bibitem[Kanade and Hebert(2012)]{kanade2012first}
Takeo Kanade and Martial Hebert.
\newblock First-person vision.
\newblock \emph{Proceedings of the IEEE}, 100\penalty0 (8):\penalty0 2442--2453, 2012.

\bibitem[Kelley(2018)]{woz_kelley}
John F.~("Jeff") Kelley.
\newblock Wizard of oz (woz): a yellow brick journey.
\newblock \emph{J. Usability Studies}, 13\penalty0 (3):\penalty0 119–124, 2018.

\bibitem[Li et~al.(2024{\natexlab{a}})Li, Ge, Ge, Wang, Wang, Zhang, and Shan]{Li_2024_CVPR}
Bohao Li, Yuying Ge, Yixiao Ge, Guangzhi Wang, Rui Wang, Ruimao Zhang, and Ying Shan.
\newblock Seed-bench: Benchmarking multimodal large language models.
\newblock In \emph{Proceedings of the IEEE/CVF Conference on Computer Vision and Pattern Recognition (CVPR)}, pages 13299--13308, 2024{\natexlab{a}}.

\bibitem[Li et~al.(2024{\natexlab{b}})Li, Zhang, Guo, Zhang, Li, Zhang, Zhang, Zhang, Li, Liu, and Li]{li2024llavaonevisioneasyvisualtask}
Bo Li, Yuanhan Zhang, Dong Guo, Renrui Zhang, Feng Li, Hao Zhang, Kaichen Zhang, Peiyuan Zhang, Yanwei Li, Ziwei Liu, and Chunyuan Li.
\newblock Llava-onevision: Easy visual task transfer, 2024{\natexlab{b}}.

\bibitem[Lin et~al.(2023)Lin, Zhu, Ye, Ning, Jin, and Yuan]{lin2023video}
Bin Lin, Bin Zhu, Yang Ye, Munan Ning, Peng Jin, and Li Yuan.
\newblock Video-llava: Learning united visual representation by alignment before projection.
\newblock \emph{arXiv preprint arXiv:2311.10122}, 2023.

\bibitem[Liu et~al.(2025)Liu, Duan, Zhang, Li, Zhang, Zhao, Yuan, Wang, He, Liu, Chen, and Lin]{mm-bench_eccv24}
Yuan Liu, Haodong Duan, Yuanhan Zhang, Bo Li, Songyang Zhang, Wangbo Zhao, Yike Yuan, Jiaqi Wang, Conghui He, Ziwei Liu, Kai Chen, and Dahua Lin.
\newblock Mmbench: Is your multi-modal model an all-around player?
\newblock In \emph{Computer Vision -- ECCV 2024}, pages 216--233, Cham, 2025. Springer Nature Switzerland.

\bibitem[Mangalam et~al.(2023)Mangalam, Akshulakov, and Malik]{Mangalam2023EgoSchemaAD}
Karttikeya Mangalam, Raiymbek Akshulakov, and Jitendra Malik.
\newblock Egoschema: A diagnostic benchmark for very long-form video language understanding.
\newblock \emph{ArXiv}, abs/2308.09126, 2023.

\bibitem[Mazzamuto et~al.(2025)Mazzamuto, Furnari, Sato, and Farinella]{Mazzamuto_2025_CVPR}
Michele Mazzamuto, Antonino Furnari, Yoichi Sato, and Giovanni~Maria Farinella.
\newblock Gazing into missteps: Leveraging eye-gaze for unsupervised mistake detection in egocentric videos of skilled human activities.
\newblock In \emph{Proceedings of the IEEE/CVF Conference on Computer Vision and Pattern Recognition (CVPR)}, pages 8310--8320, 2025.

\bibitem[Meta()]{aria_companion_app}
Meta.
\newblock Aria companion app.
\newblock \url{https://facebookresearch.github.io/projectaria_tools/docs/ARK/mobile_companion_app}.

\bibitem[Mittal et~al.(2024)Mittal, Agarwal, Lo, and Lee]{mittal2024can}
Himangi Mittal, Nakul Agarwal, Shao-Yuan Lo, and Kwonjoon Lee.
\newblock Can't make an omelette without breaking some eggs: Plausible action anticipation using large video-language models.
\newblock In \emph{Proceedings of the IEEE/CVF Conference on Computer Vision and Pattern Recognition}, pages 18580--18590, 2024.

\bibitem[Mrk{\v{s}}i{\'c} et~al.(2017)Mrk{\v{s}}i{\'c}, {\'O}~S{\'e}aghdha, Wen, Thomson, and Young]{mrksic-etal-2017-neural}
Nikola Mrk{\v{s}}i{\'c}, Diarmuid {\'O}~S{\'e}aghdha, Tsung-Hsien Wen, Blaise Thomson, and Steve Young.
\newblock Neural belief tracker: Data-driven dialogue state tracking.
\newblock In \emph{Proceedings of the 55th Annual Meeting of the Association for Computational Linguistics (Volume 1: Long Papers)}, pages 1777--1788, Vancouver, Canada, 2017. Association for Computational Linguistics.

\bibitem[Northcutt et~al.(2020)Northcutt, Zha, Lovegrove, and Newcombe]{northcutt2020egocom}
Curtis Northcutt, Shengxin Zha, Steven Lovegrove, and Richard Newcombe.
\newblock Egocom: A multi-person multi-modal egocentric communications dataset.
\newblock \emph{IEEE Transactions on Pattern Analysis and Machine Intelligence}, 2020.

\bibitem[Peddi et~al.(2024)Peddi, Arya, Challa, Pallapothula, Vyas, Gouripeddi, Wang, Zhang, Komaragiri, Ragan, Ruozzi, Xiang, and Gogate]{peddi2024captaincook4ddatasetunderstandingerrors}
Rohith Peddi, Shivvrat Arya, Bharath Challa, Likhitha Pallapothula, Akshay Vyas, Bhavya Gouripeddi, Jikai Wang, Qifan Zhang, Vasundhara Komaragiri, Eric Ragan, Nicholas Ruozzi, Yu Xiang, and Vibhav Gogate.
\newblock {CaptainCook4D: A Dataset for Understanding Errors in Procedural Activities}, 2024.

\bibitem[Plizzari et~al.(2024)Plizzari, Goletto, Furnari, Bansal, Ragusa, Farinella, Damen, and Tommasi]{plizzari2024outlook}
Chiara Plizzari, Gabriele Goletto, Antonino Furnari, Siddhant Bansal, Francesco Ragusa, Giovanni~Maria Farinella, Dima Damen, and Tatiana Tommasi.
\newblock An outlook into the future of egocentric vision.
\newblock \emph{International Journal of Computer Vision}, pages 1--57, 2024.

\bibitem[Ragusa et~al.(2023)Ragusa, Furnari, and Farinella]{ragusa_MECCANO_2023}
Francesco Ragusa, Antonino Furnari, and Giovanni~Maria Farinella.
\newblock Meccano: A multimodal egocentric dataset for humans behavior understanding in the industrial-like domain.
\newblock \emph{Computer Vision and Image Understanding (CVIU)}, 2023.

\bibitem[Ragusa et~al.(2024)Ragusa, Leonardi, Mazzamuto, Bonanno, Scavo, Furnari, and Farinella]{ragusa2023enigma51}
Francesco Ragusa, Rosario Leonardi, Michele Mazzamuto, Claudia Bonanno, Rosario Scavo, Antonino Furnari, and Giovanni~Maria Farinella.
\newblock Enigma-51: Towards a fine-grained understanding of human-object interactions in industrial scenarios.
\newblock \emph{IEEE Winter Conference on Application of Computer Vision (WACV)}, 2024.

\bibitem[Schoonbeek et~al.(2024)Schoonbeek, Houben, Onvlee, de~With, and van~der Sommen]{Schoonbeek_2024_WACV}
Tim~J. Schoonbeek, Tim Houben, Hans Onvlee, Peter~H.N. de With, and Fons van~der Sommen.
\newblock Industreal: A dataset for procedure step recognition handling execution errors in egocentric videos in an industrial-like setting.
\newblock In \emph{Proceedings of the IEEE/CVF Winter Conference on Applications of Computer Vision (WACV)}, pages 4365--4374, 2024.

\bibitem[Seminara et~al.(2024)Seminara, Farinella, and Furnari]{seminara2024differentiabletaskgraphlearning}
Luigi Seminara, Giovanni~Maria Farinella, and Antonino Furnari.
\newblock Differentiable task graph learning: Procedural activity representation and online mistake detection from egocentric videos, 2024.

\bibitem[Sener et~al.()Sener, Chatterjee, Shelepov, He, Singhania, Wang, and Yao]{sener2022assembly101}
F. Sener, D. Chatterjee, D. Shelepov, K. He, D. Singhania, R. Wang, and A. Yao.
\newblock Assembly101: A large-scale multi-view video dataset for understanding procedural activities.
\newblock \emph{CVPR 2022}.

\bibitem[Sener et~al.(2020)Sener, Singhania, and Yao]{sener2020temporal}
Fadime Sener, Dipika Singhania, and Angela Yao.
\newblock Temporal aggregate representations for long-range video understanding.
\newblock In \emph{Computer Vision--ECCV 2020: 16th European Conference, Glasgow, UK, August 23--28, 2020, Proceedings, Part XVI 16}, pages 154--171. Springer, 2020.

\bibitem[Somasundaram et~al.(2023)Somasundaram, Dong, Tang, Straub, Yan, Goesele, Engel, Nardi, and Newcombe]{Somasundaram2023ProjectAA}
Kiran~K. Somasundaram, Jing Dong, Huixuan Tang, Julian Straub, Mingfei Yan, Michael Goesele, Jakob~J. Engel, Renzo~De Nardi, and Richard~A. Newcombe.
\newblock Project aria: A new tool for egocentric multi-modal ai research.
\newblock \emph{ArXiv}, abs/2308.13561, 2023.

\bibitem[Song et~al.(2023)Song, Byrne, Nagarajan, Wang, Martin, and Torresani]{ego4d_goalstep}
Yale Song, Eugene Byrne, Tushar Nagarajan, Huiyu Wang, Miguel Martin, and Lorenzo Torresani.
\newblock Ego4d goal-step: Toward hierarchical understanding of procedural activities.
\newblock In \emph{Advances in Neural Information Processing Systems}, pages 38863--38886. Curran Associates, Inc., 2023.

\bibitem[Su et~al.(2024)Su, Ling, Shi, Cheng, Yim, and Song]{Su2024ARXIV}
Ying Su, Zhan Ling, Haochen Shi, Jiayang Cheng, Yauwai Yim, and Yangqiu Song.
\newblock {ActPlan-1K:} benchmarking the procedural planning ability of visual language models in household activities.
\newblock \emph{arXiv}, 2410.03907, 2024.

\bibitem[Team(2024{\natexlab{a}})]{Chameleon_Team_Chameleon_Mixed-Modal_Early-Fusion_2024}
Chameleon Team.
\newblock Chameleon: Mixed-modal early-fusion foundation models.
\newblock \emph{arXiv preprint arXiv:2405.09818}, 2024{\natexlab{a}}.

\bibitem[Team(2024{\natexlab{b}})]{qwen2.5}
Qwen Team.
\newblock Qwen2.5: A party of foundation models, 2024{\natexlab{b}}.

\bibitem[Tobii()]{tobiibar}
Tobii.
\newblock Tobii pro fusion bar.
\newblock \url{https://www.tobii.com/products/eye-trackers/screen-based/tobii-pro-fusion}.

\bibitem[Touvron et~al.(2023{\natexlab{a}})Touvron, Lavril, Izacard, Martinet, Lachaux, Lacroix, Rozi{\`e}re, Goyal, Hambro, Azhar, et~al.]{touvron2023llama}
Hugo Touvron, Thibaut Lavril, Gautier Izacard, Xavier Martinet, Marie-Anne Lachaux, Timoth{\'e}e Lacroix, Baptiste Rozi{\`e}re, Naman Goyal, Eric Hambro, Faisal Azhar, et~al.
\newblock Llama: Open and efficient foundation language models.
\newblock \emph{arXiv preprint arXiv:2302.13971}, 2023{\natexlab{a}}.

\bibitem[Touvron et~al.(2023{\natexlab{b}})Touvron, Martin, Stone, Albert, Almahairi, Babaei, Bashlykov, Batra, Bhargava, Bhosale, Bikel, Blecher, Ferrer, Chen, Cucurull, Esiobu, Fernandes, Fu, Fu, Fuller, Gao, Goswami, Goyal, Hartshorn, Hosseini, Hou, Inan, Kardas, Kerkez, Khabsa, Kloumann, Korenev, Koura, Lachaux, Lavril, Lee, Liskovich, Lu, Mao, Martinet, Mihaylov, Mishra, Molybog, Nie, Poulton, Reizenstein, Rungta, Saladi, Schelten, Silva, Smith, Subramanian, Tan, Tang, Taylor, Williams, Kuan, Xu, Yan, Zarov, Zhang, Fan, Kambadur, Narang, Rodriguez, Stojnic, Edunov, and Scialom]{touvron2023llama2openfoundation}
Hugo Touvron, Louis Martin, Kevin Stone, Peter Albert, Amjad Almahairi, Yasmine Babaei, Nikolay Bashlykov, Soumya Batra, Prajjwal Bhargava, Shruti Bhosale, Dan Bikel, Lukas Blecher, Cristian~Canton Ferrer, Moya Chen, Guillem Cucurull, David Esiobu, Jude Fernandes, Jeremy Fu, Wenyin Fu, Brian Fuller, Cynthia Gao, Vedanuj Goswami, Naman Goyal, Anthony Hartshorn, Saghar Hosseini, Rui Hou, Hakan Inan, Marcin Kardas, Viktor Kerkez, Madian Khabsa, Isabel Kloumann, Artem Korenev, Punit~Singh Koura, Marie-Anne Lachaux, Thibaut Lavril, Jenya Lee, Diana Liskovich, Yinghai Lu, Yuning Mao, Xavier Martinet, Todor Mihaylov, Pushkar Mishra, Igor Molybog, Yixin Nie, Andrew Poulton, Jeremy Reizenstein, Rashi Rungta, Kalyan Saladi, Alan Schelten, Ruan Silva, Eric~Michael Smith, Ranjan Subramanian, Xiaoqing~Ellen Tan, Binh Tang, Ross Taylor, Adina Williams, Jian~Xiang Kuan, Puxin Xu, Zheng Yan, Iliyan Zarov, Yuchen Zhang, Angela Fan, Melanie Kambadur, Sharan Narang, Aurelien Rodriguez, Robert Stojnic, Sergey Edunov, and Thomas
  Scialom.
\newblock Llama 2: Open foundation and fine-tuned chat models, 2023{\natexlab{b}}.

\bibitem[Vaswani(2017)]{vaswani2017attention}
A Vaswani.
\newblock Attention is all you need.
\newblock \emph{Advances in Neural Information Processing Systems}, 2017.

\bibitem[Wang et~al.(2024)Wang, Bai, Tan, Wang, Fan, Bai, Chen, Liu, Wang, Ge, Fan, Dang, Du, Ren, Men, Liu, Zhou, Zhou, and Lin]{Wang2024ARXIV}
P. Wang, S. Bai, S. Tan, S. Wang, Z. Fan, J. Bai, K. Chen, X. Liu, J. Wang, W. Ge, Y. Fan, K. Dang, M. Du, X. Ren, R. Men, D. Liu, C. Zhou, J. Zhou, and J. Lin.
\newblock {Qwen2-VL:} enhancing vision-language models perception of the world at any resolution.
\newblock \emph{arXiv}, 2409.12191, 2024.

\bibitem[Wang et~al.(2023)Wang, Kwon, Rad, Pan, Chakraborty, Andrist, Bohus, Feniello, Tekin, Frujeri, Joshi, and Pollefeys]{HoloAssist2023}
Xin Wang, Taein Kwon, Mahdi Rad, Bowen Pan, Ishani Chakraborty, Sean Andrist, Dan Bohus, Ashley Feniello, Bugra Tekin, Felipe~Vieira Frujeri, Neel Joshi, and Marc Pollefeys.
\newblock Holoassist: an egocentric human interaction dataset for interactive ai assistants in the real world.
\newblock In \emph{Proceedings of the IEEE/CVF International Conference on Computer Vision (ICCV)}, pages 20270--20281, 2023.

\bibitem[Wong et~al.(2022)Wong, Chen, Wu, Lei, Mao, Gao, and Shou]{Wong_Chen_Wu_Lei_Mao_Gao_Shou_2022}
Benita Wong, Joya Chen, You Wu, Stan~Weixian Lei, Dongxing Mao, Difei Gao, and Mike~Zheng Shou.
\newblock Assistq: Affordance-centric question-driven task completion for egocentric assistant.
\newblock \penalty0 (arXiv:2203.04203), 2022.
\newblock arXiv:2203.04203 [cs].

\bibitem[Ye et~al.(2024)Ye, Zhang, Daxberger, Chen, Lin, Li, Zhang, You, Xu, Gan, Lu, and Yang]{Ye2024ARXIV}
H. Ye, H. Zhang, E. Daxberger, L. Chen, Z. Lin, Y. Li, B. Zhang, H. You, D. Xu, Z. Gan, J. Lu, and Y. Yang.
\newblock {MM-Ego:} towards building egocentric multimodal {LLMs}.
\newblock \emph{arXiv}, 2410.07177, 2024.

\bibitem[Zhang et~al.(2022)Zhang, Wu, and Li]{zhang2022actionformer}
Chen-Lin Zhang, Jianxin Wu, and Yin Li.
\newblock Actionformer: Localizing moments of actions with transformers.
\newblock In \emph{European Conference on Computer Vision}. Springer, 2022.

\bibitem[Zhang et~al.(2024)Zhang, Wu, Li, Li, Ma, Liu, and Li]{zhang2024videoinstructiontuningsynthetic}
Yuanhan Zhang, Jinming Wu, Wei Li, Bo Li, Zejun Ma, Ziwei Liu, and Chunyuan Li.
\newblock Video instruction tuning with synthetic data, 2024.

\bibitem[Zhao et~al.(2023)Zhao, Wang, Zhang, Fu, Do, Agarwal, Lee, and Sun]{zhao2023antgpt}
Qi Zhao, Shijie Wang, Ce Zhang, Changcheng Fu, Minh~Quan Do, Nakul Agarwal, Kwonjoon Lee, and Chen Sun.
\newblock Antgpt: Can large language models help long-term action anticipation from videos?
\newblock \emph{arXiv preprint arXiv:2307.16368}, 2023.

\bibitem[Zhou et~al.(2023)Zhou, Mart{\'\i}n-Mart{\'\i}n, Kapadia, Savarese, and Niebles]{zhou2023procedure}
Honglu Zhou, Roberto Mart{\'\i}n-Mart{\'\i}n, Mubbasir Kapadia, Silvio Savarese, and Juan~Carlos Niebles.
\newblock Procedure-aware pretraining for instructional video understanding.
\newblock In \emph{Proceedings of the IEEE/CVF Conference on Computer Vision and Pattern Recognition}, pages 10727--10738, 2023.

\bibitem[Zhou et~al.(2025)Zhou, Xiao, Li, Li, Yang, Guo, Wang, Chua, and Yao]{ego-textvqa_angelayao}
Sheng Zhou, Junbin Xiao, Qingyun Li, Yicong Li, Xun Yang, Dan Guo, Meng Wang, Tat-Seng Chua, and Angela Yao.
\newblock Egotextvqa: Towards egocentric scene-text aware video question answering, 2025.

\end{thebibliography}
